\pdfoutput=1

\documentclass[11pt]{article}

\usepackage{acl}

\usepackage{times}
\usepackage{latexsym}

\usepackage[T1]{fontenc}

\usepackage[utf8]{inputenc}

\usepackage{microtype}

\usepackage{booktabs}
\usepackage{arydshln}
\usepackage{colortbl}
\usepackage{array,multirow,graphicx}
\usepackage{float}

\usepackage{graphicx}
\usepackage{pifont}
\newcommand{\cmark}{\ding{51}}%
\usepackage{wrapfig}
\usepackage{enumitem}
\usepackage{makecell}
\usepackage{tikz}
\usepackage{collcell}
\usepackage{diagbox}
\usepackage{rotating}
\usepackage{pgf}
\usepackage{transparent}
\usepackage{subcaption}

\usepackage{amsmath,amsfonts,bm}
\usepackage[nameinlink]{cleveref}

\crefformat{section}{\S#2#1#3} 
\crefname{algorithm}{Alg.}{Algs.}
\crefformat{subsection}{\S#2#1#3}
\Crefname{equation}{Eq.}{Eqs.}
\Crefname{figure}{Fig.}{Figs.}

\makeatletter   
\newcommand{\sveryshortarrow}[1][3pt]{\mathrel{%
    \vcenter{\hbox{\rule[-.5\fontdimen8\scriptfont3]
               {\scriptratio\dimexpr#1\relax}{\fontdimen8\scriptfont3}}}%
   \mkern-4mu\hbox{\let\f@size\sf@size\usefont{U}{lasy}{m}{n}\symbol{41}}}}
\makeatother

\def\eqref#1{equation~\ref{#1}}

\def\1{\bm{1}}

\def\vx{{\bm{x}}}
\def\vy{{\bm{y}}}

\def\m1{{\bm{1}}}

\DeclareMathAlphabet{\mathsfit}{\encodingdefault}{\sfdefault}{m}{sl}
\SetMathAlphabet{\mathsfit}{bold}{\encodingdefault}{\sfdefault}{bx}{n}

\title{On Context Utilization in Summarization with Large Language Models}

\author{Mathieu Ravaut$^1$$^,$$^2$,~
Aixin Sun$^1$,
Nancy F. Chen$^{1,2,4,5}$, Shafiq Joty$^{1,3}$\\
$^1$ Nanyang Technological University, Singapore\\
$^2$ Institute of Infocomm Research (I$^{2}$R), A$^{*}$STAR, Singapore\\
$^3$ Salesforce Research\\
$^4$ CNRS@CREATE, Singapore\\
$^5$ Centre for Frontier AI Research (CFAR), A*STAR, Singapore\\
\texttt{\{mathieuj001@e.ntu, axsun@ntu, srjoty@ntu\}.edu.sg}\\
\texttt{nfychen@i2r.a-star.edu.sg}
}

\begin{document}

\maketitle

\begin{abstract}
Large language models (LLMs) excel in abstractive summarization tasks, delivering fluent and pertinent summaries. Recent advancements have extended their capabilities to handle long-input contexts, exceeding 100k tokens. However, in question answering, language models exhibit uneven utilization of their input context. They tend to favor the initial and final segments, resulting in a U-shaped performance pattern concerning where the answer is located within the input. This bias raises concerns, particularly in summarization where crucial content may be dispersed throughout the source document(s). Besides, in summarization, mapping facts from the source to the summary is not trivial as salient content is usually re-phrased. In this paper, we conduct the first comprehensive study on context utilization and position bias in summarization. Our analysis encompasses 6 LLMs, 10 datasets, and 5 evaluation metrics. We introduce a new evaluation benchmark called MiddleSum on the which we benchmark two alternative inference methods to alleviate position bias: hierarchical summarization and incremental summarization\footnote{Our code and data can be found here: \url{https://github.com/ntunlp/MiddleSum}.}. 

\end{abstract}

\section{Introduction}
\label{sec:intro}
Large language models (LLMs) have drastically transformed the landscape of NLP recently \cite{gpt3}. 
With instruction tuning \cite{instructgpt,flant5}, LLMs made a major leap forward in conditional (prompted) content generation, and can generate satisfying outputs without the need to fine-tune on a specific task.
In abstractive summarization specifically, this approach has arguably opened a new paradigm: summaries generated by LLMs are highly fluent, grammatical and relevant \citep{goyal2022news}. 
Despite noticeably lower scores on automatic metrics such as ROUGE \citep{rouge} or BERTScore \citep{bertscore}, summaries generated by LLMs are largely preferred by humans over summaries from state-of-the-art fine-tuned models like BRIO \citep{brio,liu2023towards}. 
In fact, on XSum, GPT-3.5 summaries are even on par with re-annotated human-written summaries, and much better than the dataset's original ground-truth, according to human evaluators \citep{zhang2023benchmarking}. 
LLMs also show promising capability in evaluating summaries generated by other systems, including LLMs \citep{gptscore, luo2023chatgpt, shen2023large}.


Despite this success, a few major technological bottlenecks remain with LLMs, including the maximum length of their context window. 
The standard context window length for open-source LLMs is 2k tokens \citep{gpt3, bloom, falcon, llama}, which drastically limits their usefulness for long-input summarization \citep{scrolls}. 
Several techniques were proposed to extend the context window, including ALiBi \citep{alibi}, LeX \citep{lex}, position interpolation \citep{posinterp} and YaRN \citep{yarn}. 
While some of them claim up to 100k+ tokens processing capacity \citep{yarn}, it remains unclear how much such methods help on long-context summarization. 

Scaling up context length would only succeed if a key question gets addressed first: \textbf{\emph{do LLMs make proper use of their entire context?}} 
Recent work \citep{litm} suggested that, surprisingly, such a simple assumption may not hold: through experiments on multi-document question answering and key-value retrieval, the authors find that LLMs mostly focus on the \emph{beginning} and \emph{end} of the (long) context window. 
Plotting performance with regards to the position of the important information exhibits a \emph{U-shape}, with performance high at first (beginning of the source), then dropping, and rising again at the end. 
Worryingly, in the middle of the context window, LLMs' performance can drop to even \emph{below random chance}, calling for greater examination of LLMs' behaviors with regard to the position of information within the source. 

In this work, we investigate in depth how LLMs use their context window in abstractive summarization. 
Unlike in question-answering, mapping facts in the output to a specific snippet in the source is not straightforward in abstractive summarization, due to the high-level of re-phrasing and compression.
We conduct a large-scale study with 6 LLMs, 10 datasets covering many aspects of summarization, and 5 highly diverse automatic metrics.
Our contributions are threefold:
\begin{itemize}[leftmargin=*,itemsep=0.1em]
    \item We conduct the first large-scale analysis on context utilization in abstractive summarization, and the impact of the position of salient information on performance. We show that the U-shape or \emph{middle-curse} exhibited by \citep{litm} also holds in abstractive summarization.

    \item We craft an evaluation dataset (\textbf{MiddleSum}) where important information is concentrated in the middle of the context, enabling us to automatically quantify how much LLMs are affected by the \emph{middle-curse}.

    \item We benchmark two alternative methods for inference on MiddleSum: \emph{hierarchical summarization} and \emph{incremental summarization}, showing their promise at alleviating the \emph{middle curse} (especially in the scientific paper domain). 
    
\end{itemize}

\section{Experimental Setup}
\label{sec:setup}
\paragraph{Datasets}
We cover a broad set of diverse abstractive summarization tasks, varying length and domain. We include 5 datasets of standard length (source is below 2k tokens, which always fits in the  context window): (i) \textbf{CNN/DailyMail} \citep{cnndm}, (ii) \textbf{XSum} \citep{xsum}, (iii) \textbf{Reddit-TIFU} \citep{reddit}, (iv) \textbf{SAMSum} \citep{samsum}, and (v) \textbf{Multi-XScience} \citep{multixscience}. We also include another 5 long-input summarization datasets: (i) \textbf{Arxiv} and (ii) \textbf{PubMed} \citep{arxivpubmed}, (iii) \textbf{GovReport} \citep{govreport}, (iv) \textbf{SummScreenFD} \citep{summscreen}, and (v) \textbf{Multi-News} \citep{multinews}. A high-level view of each dataset is shown in \Cref{datasets}, and detailed statistics are presented in \Cref{sec:appendix_a}. For all datasets, we run experiments on the test set, subsampling 1,000 data points if its size is greater than 1,000, or using the entire test set otherwise. 


\begin{figure*}
    \centering 
    \includegraphics[width=\textwidth]{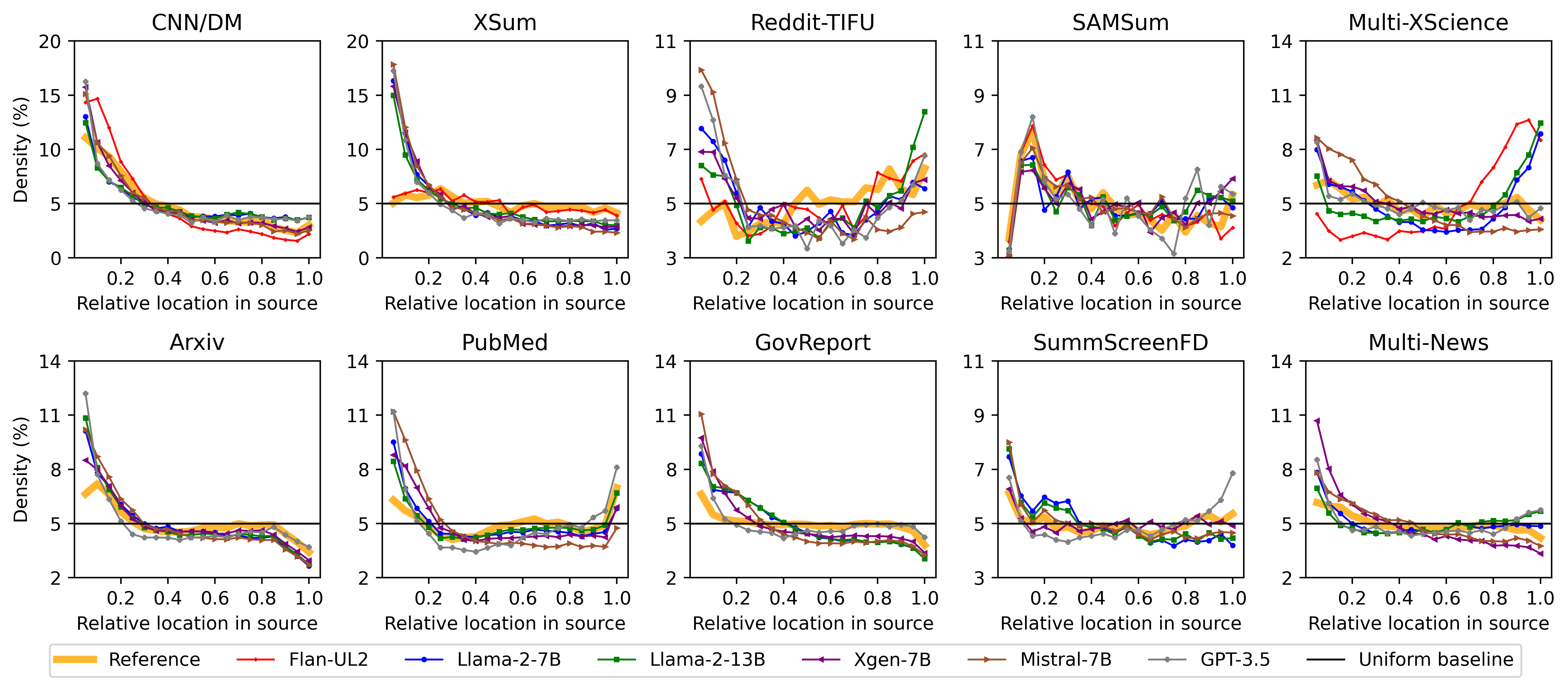}
    \caption{\small Distribution of the relative location of summary bigrams within the source. We split each source document into 20 bins of the same number of words, and plot the distribution of summary bigrams over source bins.}
    \label{alignment_absolute}
\end{figure*}

\paragraph{Models}
We experiment with 6 popular and high-performing LLMs:
\begin{itemize}[leftmargin=*,itemsep=0.1em]
    \item \textbf{Flan-UL2} is a 20B parameters encoder-decoder model pre-trained on 1T tokens. It is based on the UL2 20B model \citep{ul2}, with the addition of Flan-T5 \citep{flant5} instruction fine-tuning. The context window is 2k tokens.
    
    \item \textbf{Llama-2} \citep{llama2} is a recently introduced powerful decoder-only model pre-trained on 2T tokens, ranging from 7B to 70B parameters, and with a 4k tokens context window. We use the \textbf{7B} and \textbf{13B} models. 
    
    \item \textbf{Xgen-7B} \citep{xgen} is a 7B decoder-only model pre-trained on up to 1.5T tokens. It supports an 8k tokens context window. 

    \item \textbf{Mistral-7B} \citep{mistral} is also an 8k-context 7B decoder-only model, with performance slightly better than Llama-2-13B. 

    \item \textbf{GPT-3.5} (\texttt{gpt-3.5-turbo-0125}), which is known to be good at generating summaries and has a 16k tokens context window.\footnote{To reduce cost, we subsample 300 data points when using GPT-3.5. Our total API cost is inferior to 300 USD.}
\end{itemize}

\begin{table}[]
\centering
\setlength{\tabcolsep}{2pt}
\resizebox{1.0\columnwidth}{!}{
\begin{tabular}{l|cc|cc|cc}

\toprule 

& \multicolumn{2}{c|}{\textbf{Input length}} 
& \multicolumn{2}{c|}{\textbf{\# Documents}} 
& \multicolumn{2}{c}{\textbf{In Flan?}} \\
\textbf{Dataset} 
& Standard          
& Long  
& Single          
& Multi          
& Yes
& No \\

\midrule

CNN/DM \citep{cnndm}                 & \cmark & & \cmark & & \cmark & \\
XSum \citep{xsum}                    & \cmark & & \cmark & & \cmark & \\
Reddit-TIFU \citep{reddit}           & \cmark & & \cmark & & & \cmark \\
SAMSum \citep{samsum}                & \cmark & & \cmark & & \cmark & \\
Multi-XScience \citep{multixscience} & \cmark & & & \cmark & & \cmark \\  

\cdashline{1-7}

Arxiv \citep{arxivpubmed}            & & \cmark & \cmark & & & \cmark \\
PubMed \citep{arxivpubmed}           & & \cmark & \cmark & & & \cmark \\
GovReport \citep{govreport}          & & \cmark & \cmark & & & \cmark \\
SummScreenFD \citep{summscreen}        & & \cmark & \cmark & & & \cmark \\
Multi-News \citep{multinews}         & & \cmark & & \cmark & \cmark & \\

\bottomrule

\end{tabular}
}
\caption{\small Summarization datasets under study. In standard length datasets, the context and summary fit within a 2k tokens LLM context window.}
\vspace{-1.0em}
\label{datasets}
\end{table}

We analyze the open-source models through HuggingFace \emph{transformers} library \citep{transformers}, and use the OpenAI API for GPT-3.5. 
We use the instruction-tuned (or \emph{chat}) checkpoints for Llama-2, Xgen-7B and Mistral-7B. Note that popular instruction-tuning datasets such as Flan \citep{flan} include some of the datasets we study: CNN/DM, XSum, SAMSum and Multi-News. 
To run inference, we use the following prompt: \texttt{Read the following text and summarize it: [text]. Summarize the above text in [n] sentences. Summary:} where \texttt{n} is set to an average number of target sentences per dataset (see \Cref{sec:appendix_a}). We infer all models in \emph{bfloat16} and sample summaries with top-k sampling \citep{topk} using $k=50$ and temperature $T=0.3$.

\paragraph{Evaluation Measures}
Summarization evaluation is especially challenging in the LLM era, as most automatic metrics poorly correlate with human preferences \citep{goyal2022news,liu2022revisiting}. 
To get a broad picture of performance, we evaluate with metrics as diverse as possible. 
First, we consider \textbf{reference-based} metrics: \textbf{ROUGE-2} \citep{rouge}, which measures bigram overlap, \textbf{BERTScore} \citep{bertscore}, which measures semantic similarity with BERT \citep{bert} embeddings, and \textbf{A3CU} \citep{liu2023towards}, which extracts facts in the form of Atomic Content Units (ACUs) \cite{liu2022revisiting}, and checks the presence of ACUs between prediction and reference. 
As \textbf{reference-free} metrics, we include \textbf{SummaC} \citep{summac}, a leading factual consistency evaluation metric relying on entailment scores between pairs of source and summary sentences. 
We also leverage \textbf{GPT-3.5} again (still \texttt{gpt-3.5-turbo-0125}), this time as a summarization \emph{evaluator}, which is proven to be a strong natural language generation evaluator \citep{wang2023chatgpt, shen2023large, jain-etal-2023-multi}. We prompt the model with the source and generated summary (which fits in GPT-3.5's 16k context window) and ask to output a score on a likert scale from 1 to 5. We refer to \Cref{sec:appendix_b} for the full prompt template.\footnote{We also subsample 300 data points when using GPT-3.5 as an evaluator, to reduce API cost.}

We report the performance of LLMs on all 10 datasets, alongside a comparison to SOTA, in \Cref{sec:appendix_c}. FLan-UL2 dominates on standard-length datasets, but GPT-3.5 has the upper hand on the long-input ones. Performance itself is not our focus in this paper, but rather which position-related factors influence it. We discard Flan-UL2 on long-input datasets due to very poor performance.


\begin{figure*}[t]
    \centering 
    \includegraphics[width=\textwidth]{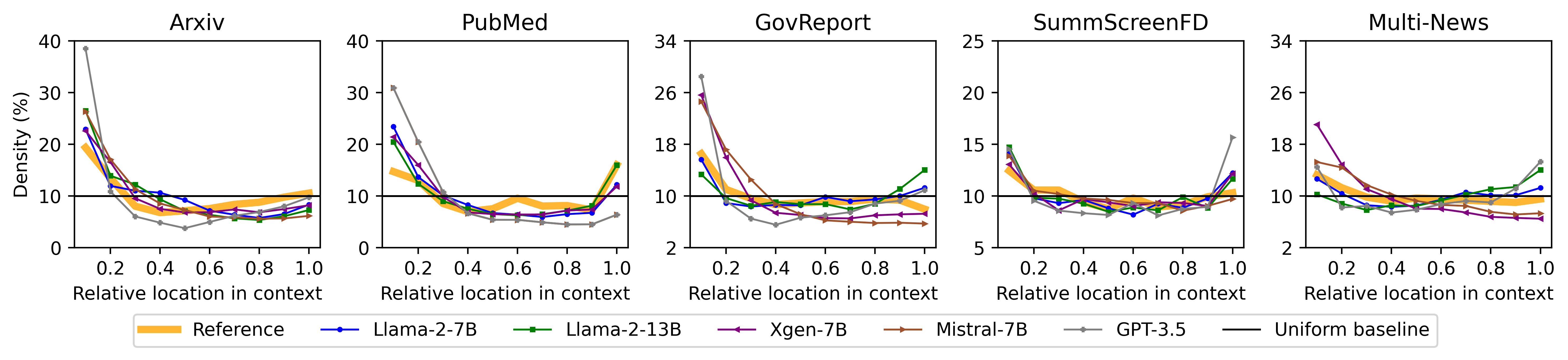}
    \caption{\small Distribution of relative location of input context sentences aligned with sentences from summaries. X-axis corresponds to the source sentence bin, y-axis to the fraction of aligned sentences in each bin.}
    \label{alignment_relative}
\end{figure*}

\section{Experiments}
\label{sec:exps}
In this section, we describe a series of experiments aimed at understanding how LLMs treat information in their input depending on the position. 

\subsection{RQ1: Where in the \emph{source} do LLMs take their information from?}
\label{subsec:rq1}

We first investigate summaries generated by LLMs, and map them to specific parts of the input. Unlike in question-answering or \emph{extractive} summarization, mapping salient information from a summary to the source is not trivial in \emph{abstractive} summarization. 

We follow the approach used in \citep{reddit, narrasum} and compute the relative position of bigams from generated summaries within the source documents, as a proxy for the position of salient information. We only use unique bigrams from summaries, and for each bigram, find all its occurrences within the source, if there are any. We then split the source into 20 bins of the same number of words, and compute the fraction of matched bigrams found in each bin. On top of the LLMs described above, we include the position of bigrams from reference summaries, and a uniform baseline. 

As seen in \Cref{alignment_absolute}, all summarization datasets except XSum, and Reddit-TIFU show some \emph{lead bias}: salient bigrams from the reference (orange curves) are more likely to be found at the beginning of the source. However, LLMs show a significantly stronger lead bias on all datasets: bigrams from LLMs summaries are much more likely to be found in the first 20\% words of the source. It is especially striking on XSum (except for Flan-UL2), Reddit-TIFU, Arxiv, PubMed and GovReport. On XSum, Flan-UL2 closely matches the reference distribution, which we attribute to its better instruction tuning. Results in \Cref{sec:appendix_d} confirm that bigram distribution for LLMs and references are statistically different (p-value of Kolmogorov-Smirnov test \citep{ks} inferior to 0.001) in all but 4 out of the 55 (dataset, LLM) setups: Flan-UL2 on XSum and SAMSum, Llama-2-7B on SAMSum and GPT-3.5 on SAMSum. We conclude that \textbf{\emph{LLMs focus on contents at the beginning of the source document(s).}}

\subsection{RQ2: Where do LLMs look at within their \emph{context window}?}
\label{subsec:rq2}

\begin{table*}[]
\centering

\resizebox{\textwidth}{!}{
\begin{tabular}{l@{\hspace{10pt}}l@{\hspace{10pt}}cccccc|cccccc}

\toprule 

\textbf{Metric}                        
& \textbf{Model} 
& \multicolumn{1}{c}{\rotatebox[origin=l]{0}{\textbf{CNN/DM}}}
& \multicolumn{1}{c}{\rotatebox[origin=l]{0}{\textbf{XSum}}}
& \multicolumn{1}{c}{\rotatebox[origin=l]{0}{\textbf{Reddit}}} 
& \multicolumn{1}{c}{\rotatebox[origin=l]{0}{\textbf{SAMSum}}} 
& \multicolumn{1}{c}{\rotatebox[origin=l]{0}{\textbf{Multi-X}}} 
& \multicolumn{1}{c}{\rotatebox[origin=l]{0}{\textbf{\emph{AVG}}}} 
& \multicolumn{1}{c}{\rotatebox[origin=l]{0}{\textbf{Arxiv}}} 
& \multicolumn{1}{c}{\rotatebox[origin=l]{0}{\textbf{PubMed}}} 
& \multicolumn{1}{c}{\rotatebox[origin=l]{0}{\textbf{GovReport}}} 
& \multicolumn{1}{c}{\rotatebox[origin=l]{0}{\textbf{SummScreenFD}}} 
& \multicolumn{1}{c}{\rotatebox[origin=l]{0}{\textbf{Multi-N}}}  
& \multicolumn{1}{c}{\rotatebox[origin=l]{0}{\textbf{\emph{AVG}}}} \\

\midrule 

\multirow{6}{*}{\textbf{ROUGE-2}}
& Flan-UL2
& -0.296 & -0.124 & \transparent{0.5}0.048 & -0.069 & -0.201 & \emph{-0.128}
& \_ & \_ & \_ & \_ & \_ & \_ \\
& Llama-2-7B 
& -0.160 & \transparent{0.5}-0.023 & 0.063 & \transparent{0.5}-0.059 & -0.100 & \emph{-0.056} 
& \transparent{0.5}0.022 & -0.113 & \transparent{0.5}-0.109 & \transparent{0.5}-0.079 & -0.210 & \emph{-0.098} \\
& Llama-2-13B
& -0.166 & -0.086 & \transparent{0.5}0.031 & -0.078 & \transparent{0.5}-0.039 & \emph{-0.068} 
& \transparent{0.5}-0.017 & -0.081 & -0.166 & \transparent{0.5}-0.139 & -0.213 & \emph{-0.123} \\
& Xgen-7B
& -0.228 & \transparent{0.5}-0.042 & 0.066 & \transparent{0.5}-0.039 & \transparent{0.5}-0.041 & \emph{-0.056} 
& \transparent{0.5}0.028 & -0.091 & -0.405 & \transparent{0.5}0.063 & -0.283 & \emph{-0.138} \\ 
& Mistral-7B   
& -0.289 & \transparent{0.5}-0.031 & \transparent{0.5}0.006 & \transparent{0.5}-0.024 & \transparent{0.5}-0.052 & \emph{-0.078} 
& -0.270 & \transparent{0.5}-0.279 & -0.585 & -0.132 & -0.324 & \emph{-0.318} \\ 
& GPT-3.5 
& -0.323 & \transparent{0.5}-0.027 & \transparent{0.5}-0.031 & \transparent{0.5}-0.097 & \transparent{0.5}0.088 & \emph{-0.078} 
& \transparent{0.5}0.026 & \transparent{0.5}-0.093 & -0.123 & \transparent{0.5}-0.061 & -0.233 & \emph{-0.097} \\

\cdashline{1-14}

\multirow{6}{*}{\textbf{BERTScore}}
& Flan-UL2
& -0.331 & -0.185 & \transparent{0.5}0.062 & -0.144 & -0.399 & \emph{-0.187}
& \_ & \_ & \_ & \_ & \_ & \_ \\
& Llama-2-7B 
& -0.173 & \transparent{0.5}-0.012 & \transparent{0.5}0.062 & -0.130 & -0.385 & \emph{-0.128} 
& \transparent{0.5}-0.031 & -0.203 & \transparent{0.5}-0.104 & \transparent{0.5}-0.067 & -0.256 & \emph{-0.132} \\
& Llama-2-13B
& -0.193 & -0.102 & \transparent{0.5}0.038 & -0.089 & -0.352 & \emph{-0.140} 
& \transparent{0.5}-0.082 & -0.209 & \transparent{0.5}-0.063 & \transparent{0.5}-0.152 & -0.279 & \emph{-0.157} \\
& Xgen-7B
& -0.252 & -0.106 & \transparent{0.5}0.046 & -0.075 & -0.343 & \emph{-0.146} 
& \transparent{0.5}-0.017 & -0.125 & -0.353 & \transparent{0.5}-0.093 & -0.345 & \emph{-0.187} \\
& Mistral-7B   
& -0.278 & -0.052 & \transparent{0.5}0.014 & -0.108 & -0.416 & \emph{-0.168} 
& -0.348 & -0.367 & -0.567 & -0.356 & -0.403 & \emph{-0.408} \\
& GPT-3.5 
& -0.280 & \transparent{0.5}-0.021 & \transparent{0.5}-0.025  & -0.174 & -0.247 & \emph{-0.149} 
& \transparent{0.5}-0.087 & -0.210 & -0.175 & -0.142 & -0.262 & \emph{-0.175} \\

\cdashline{1-14}

\multirow{6}{*}{\textbf{A3CU}}
& Flan-UL2
& -0.258 & -0.090 & \transparent{0.5}0.050 & -0.123 & -0.069 & \emph{-0.098}
& \_ & \_ & \_ & \_ & \_ & \_ \\
& Llama-2-7B 
& -0.182 & -0.076 & \transparent{0.5}0.028 & -0.121 & -0.090 & \emph{-0.088} 
& \transparent{0.5}-0.038 & -0.209 & -0.154 & \transparent{0.5}-0.129 & -0.217 & \emph{-0.149} \\ 
& Llama-2-13B
& -0.190 & -0.098 & \transparent{0.5}0.009 & -0.166 & \transparent{0.5}0.016 & \emph{-0.086} 
& -0.104 & -0.232 & \transparent{0.5}-0.111 & -0.198 & -0.228 & \emph{-0.175} \\ 
& Xgen-7B
& -0.212 & -0.130 & \transparent{0.5}0.023 & -0.126 & \transparent{0.5}-0.025 & \emph{-0.094} 
& \transparent{0.5}-0.036 & -0.255 & -0.211 & \transparent{0.5}-0.076 & -0.287 & \emph{-0.173} \\ 
& Mistral-7B   
& -0.291 & -0.110 & \transparent{0.5}\-0.004 & -0.105 & -0.119 & \emph{-0.126} 
& -0.160 & -0.283 & -0.305 & \transparent{0.5}-0.010 & -0.283 & \emph{-0.208} \\ 
& GPT-3.5 
& -0.256 & \transparent{0.5}-0.022 & \transparent{0.5}-0.039 & -0.216 & 0.141 & \emph{-0.078} 
& -0.064 & -0.293 & -0.201 & \transparent{0.5}-0.059 & -0.264 & \emph{-0.176} \\

\midrule 

\multirow{6}{*}{\textbf{SummaC}}
& Flan-UL2
& \transparent{0.5}-0.012 & 0.548 & 0.270 & 0.186 & \transparent{0.5}-0.035 & \emph{0.191}
& \_ & \_ & \_ & \_ & \_ & \_ \\
& Llama-2-7B 
& 0.088 & 0.552 & 0.375 & 0.227 & 0.224 & \emph{0.293} 
& 0.090 & 0.108 & 0.126 & \transparent{0.5}-0.020 & 0.205 & \emph{0.102} \\ 
& Llama-2-13B
& 0.162 & 0.556 & 0.394 & 0.173 & 0.096 & \emph{0.276} 
& 0.090 & 0.265 & 0.192 & \transparent{0.5}-0.144 & 0.232 & \emph{0.127} \\ 
& Xgen-7B
& \transparent{0.5}0.001 & 0.161 & 0.220 & 0.117 & \transparent{0.5}0.004 & \emph{0.101}
& -0.208 & -0.087 & -0.313 & -0.141 & \transparent{0.5}0.046 & \emph{-0.141} \\ 
& Mistral-7B   
& \transparent{0.5}-0.045 & 0.515 & 0.149 & 0.069 & 0.154 & \emph{0.128} 
& -0.250 & -0.103 & -0.387 & 0.124 & \transparent{0.5}-0.010 & \emph{-0.125} \\ 
& GPT-3.5   
& 0.156 & 0.590 & 0.444 & 0.180 & \transparent{0.5}0.008 & \emph{0.276} 
& \transparent{0.5}0.058 & 0.237 & \transparent{0.5}0.061 & \transparent{0.5}-0.055 & \transparent{0.5}0.089 & \emph{0.078} \\ 

\cdashline{1-14}

\multirow{6}{*}{\textbf{GPT-3.5}}
& Flan-UL2
& \transparent{0.5}-0.020 & 0.196 & \transparent{0.5}0.027 & \transparent{0.5}-0.009 & -0.193  & \emph{0.000} 
& \_ & \_ & \_ & \_ & \_ & \_ \\
& Llama-2-7B 
& \transparent{0.5}0.036 & \transparent{0.5}0.036 & -0.152 & \transparent{0.5}0.077 & -0.153 & \emph{-0.031} 
& \transparent{0.5}0.008 & \transparent{0.5}-0.116 & \transparent{0.5}-0.013 & \transparent{0.5}-0.068 & -0.120 & \emph{-0.062} \\
& Llama-2-13B
& \transparent{0.5}0.039 & \transparent{0.5}0.072 & \transparent{0.5}-0.038 & \transparent{0.5}0.066 & \transparent{0.5}-0.052 & \emph{0.017} & \transparent{0.5}-0.010 & \transparent{0.5}-0.084 & \transparent{0.5}-0.084 & \transparent{0.5}-0.060 & \transparent{0.5}-0.051 & \emph{-0.058} \\
& Xgen-7B
& \transparent{0.5}0.056 & \transparent{0.5}-0.007 & \transparent{0.5}-0.101 & \transparent{0.5}0.006 & -0.174 & \emph{-0.044} 
& \transparent{0.5}-0.058 & \transparent{0.5}-0.096 & -0.317 & \transparent{0.5}-0.063 & \transparent{0.5}-0.055  &\emph{-0.118} \\
& Mistral-7B   
& -0.115 & 0.124 & -0.133 & \transparent{0.5}0.036 & -0.204 & \emph{-0.058} 
& -0.446 & -0.322 & -0.580 & -0.188 & -0.163  & \emph{-0.342} \\
& GPT-3.5   
& \transparent{0.5}0.024 & \transparent{0.5}0.014 & \transparent{0.5}-0.108 & 0.131 & \transparent{0.5}-0.054 & \emph{0.001} 
& 0.156 & \transparent{0.5}-0.008 & -0.172 & \transparent{0.5}0.068 & \transparent{0.5}0.024 & \emph{0.014} \\

\bottomrule

\end{tabular}
}
\vspace{-0.5em}
\caption{\small Spearman correlation coefficient between each LLM's metric, and the mean position of salient information within the context window. Flan-UL2 is not applied to long-context summarization datasets due to its too short context window. \textbf{Multi-X} is short for Multi-XScience, \textbf{Multi-N} is Multi-News dataset, \textbf{AVG} columns represent the average over standard-length and long-input datasets, respectively. Numbers in gray correspond to non-significant Spearman scores (p-value greater than 0.05).}
\label{corr_pos}
\vspace{-0.5em}
\end{table*}

In the previous experiment's design, LLMs may not see the entire source in long-input summarization datasets, due to their limited context window, which is shorter than the source on the long-input datasets. We now focus on input information accessible to LLMs, and only consider salient information if it falls within the context window. Besides, since the same bigram may occur multiple times throughout the source, we adjust the methodology for saliency estimation. We align sentences in generated summaries to sentences in the context, following the procedure described in \citep{neusum} and also used in \citep{cod}. Specifically, we greedily select source sentences (among the ones fitting in context window) until the ROUGE-1 F1 score between the set of selected source sentences and the summary stops increasing. The resulting set of source sentences forms a proxy of the visible salient input information being rephrased by the model when summarizing. We split each truncated source document into 10 bins of the same number of sentences, and map each aligned source sentence to its bin. Note that bins are not directly comparable across models, as context length varies across models.

As we can see in \Cref{alignment_relative}, sentences from the first 10\% or last 10\% of the input context are much more represented than others. A clear U-shape emerges on PubMed and SummScreenFD for all LLMs. This is intriguing knowing that the LLMs have different context window lengths, and the last 10\% of each context window may contain content of varying saliency. In other words, \emph{\textbf{LLMs seem to be mostly re-phrasing information from the beginning or the end of their context window}}. 

Results in \Cref{sec:appendix_e} confirm an even stronger position bias with \emph{base} models.

\subsection{RQ3: Does LLMs performance depend on the position of salient information?}
\label{subsec:rq3}

Results from the last experiment raise the question of whether LLMs' summarization performance changes depending on \emph{where salient information is located within the input}. As an approximation for salient information, we consider the alignment between summary sentences and source sentences like in \Cref{alignment_relative}, but this time using the \emph{reference} summaries. Each reference summary is mapped to the source sentences it maximizes ROUGE-1 F1 against, which may be scattered across the whole source. We convert each source sentence to its cumulative word count from the beginning of the source, and take the average as an approximation of the mean position of salient information within the source. We keep data points with mean salient position fitting within the LLM context window.

We examine performance changes with regard to this salient position. To do so, we compute the Spearman correlation coefficient between salient position and each evaluation metric in \Cref{corr_pos}. A high absolute Spearman value means that summary quality (as measured by this metric) can change (and deteriorate) with the position of important information within the context. 

There are several takeaway findings from this Table. First, we notice that on standard-length datasets, reference-based evaluation metrics are negatively correlated to position of salient information. The correlation is only moderate, yet remarkably consistent across datasets (except Reddit-TIFU) and models. This is surprising, since such datasets fit entirely in context and are not affected by truncation. In contrast, reference-free metrics show either no significant or \emph{positive} correlation to information position. For long-input datasets, the negative trend for reference-based metrics is confirmed. On these lengthy datasets, SummaC and GPT-3.5 tend to switch from positive to negative correlation, especially for Xgen-7B and Mistral-7B. We highlight that since GPT-3.5 itself is affected by the \emph{middle-curse} from \citet{litm}, it may not accurately evaluate summarization when salient content lays in the middle of the context. In light of these results, we conservatively conclude that \emph{\textbf{LLMs' summarization performance is sensitive to the position of salient information in the context window}}.

\section{Analysis}
\label{sec:rw}

\begin{figure*}[t]
    \centering 
    \includegraphics[width=\textwidth]{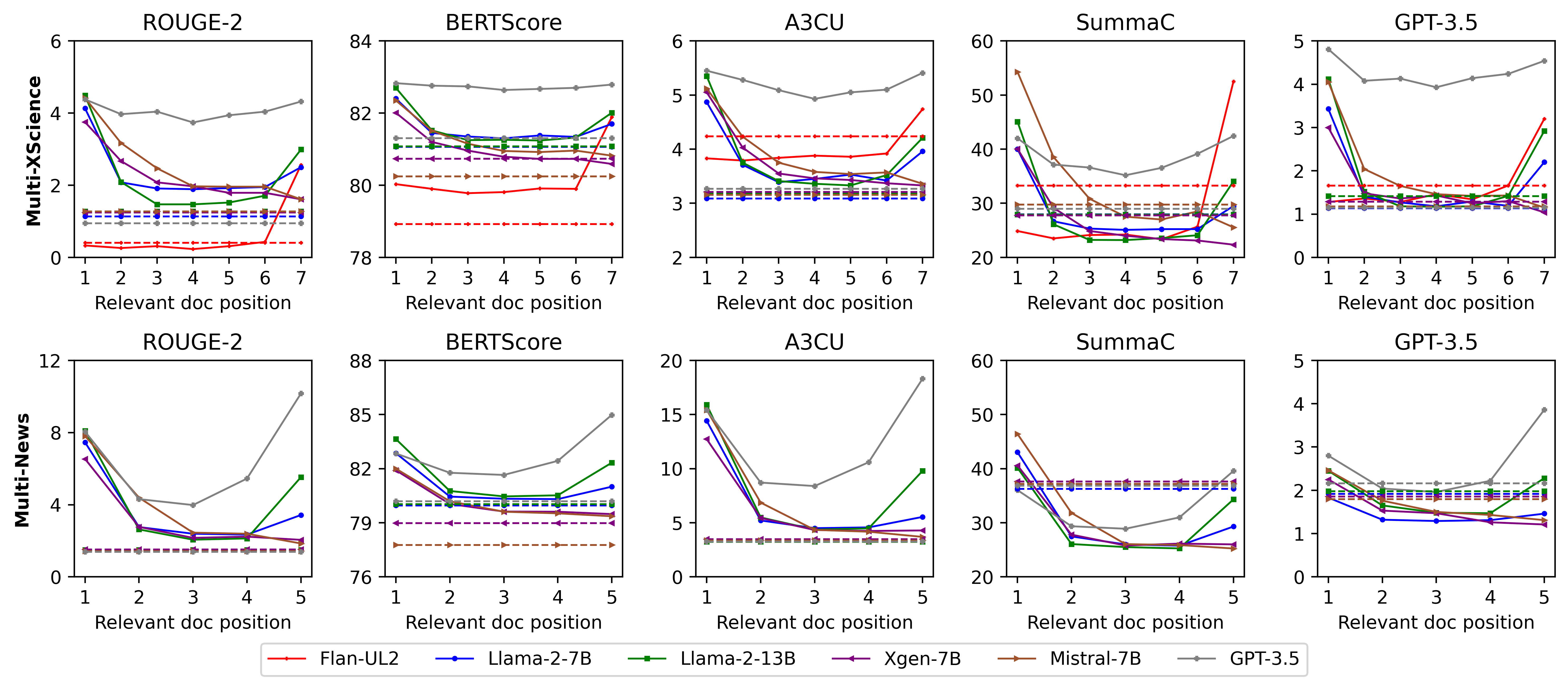}
    \caption{\small Multi-document summarization performance on Multi-XScience (top row) and Multi-News (bottom row) when a unique relevant document is used, and its position is varied (x-axis). Dashed horizontal lines correspond to the random baseline.}
    \label{mds_salient}
\end{figure*}


\begin{figure}[h]
    \centering 
    \includegraphics[width=\columnwidth]{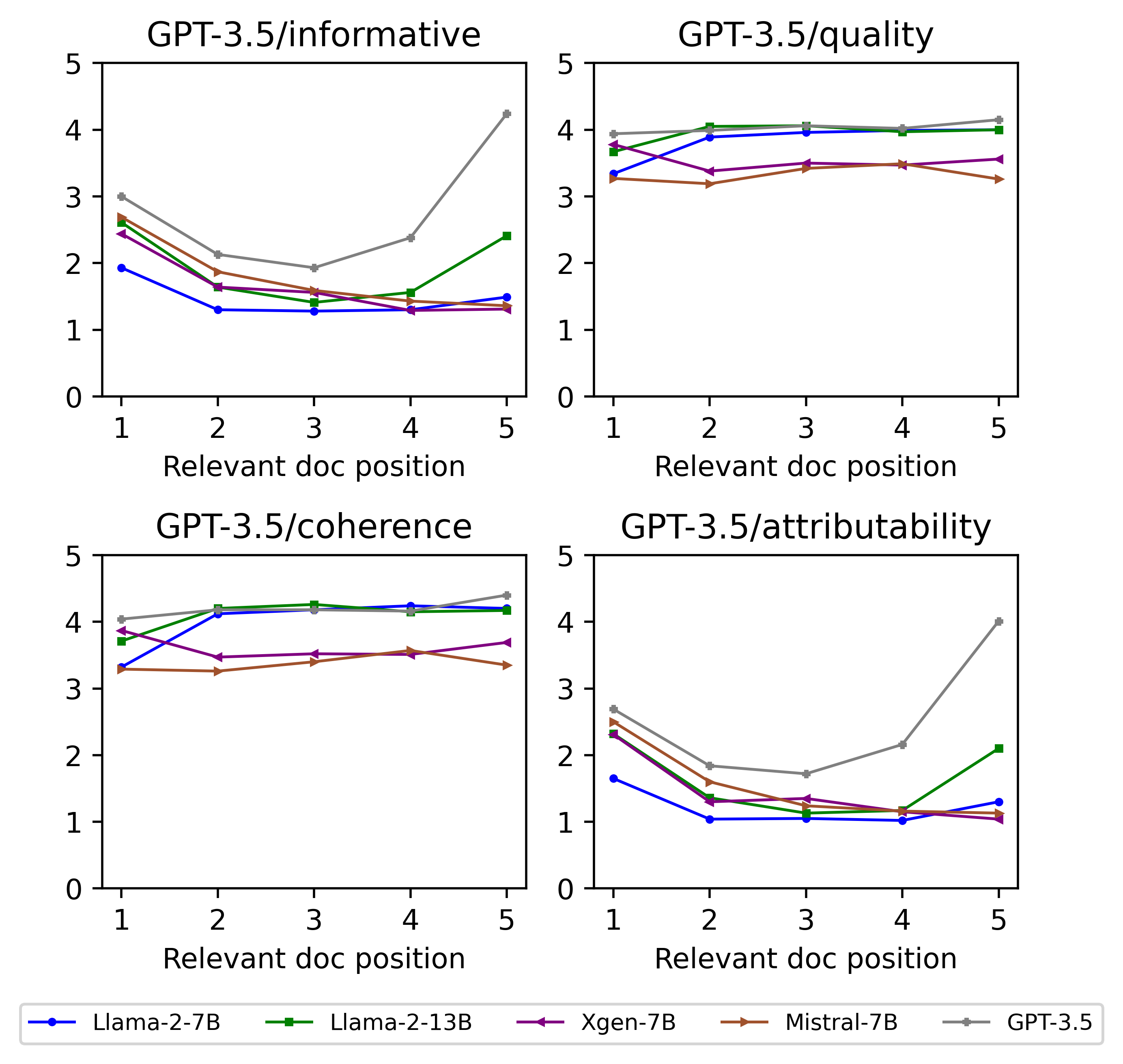}
    \caption{\small Fine-grained evaluation of multi-document summarization on Multi-News with GPT-3.5 when varying the position of a unique relevant input document.}
    \label{mds_salient_gpt}
\end{figure}

\subsection{How is information in the \emph{middle} treated?}
\label{subsec:4_1}

Previous experiments show that LLMs place more emphasis on the beginning and the end of their context. We now narrow down on how LLMs treat the \emph{middle}.
To remove the effect of spread of salient information, we perform two controlled  experiments in multi-document summarization. This setup enables us to shuffle the order of the input, which is not realistic for the single-document setup as it would break coherence. We only consider data points with the same number $k$ of documents: $k=7$ documents for Multi-XScience ($n=329$), and $k=5$ documents for Multi-News ($n=219$)\footnote{We don't subsample from these subsets for GPT-3.5.}.

In the first experiment, we vary the position of salient information throughout the input. We keep a single document (the abstract of the query paper on Multi-XScience, and the document with the highest BERTScore with the reference on Multi-News), and place it at position $j$ for $j \in \{1,\ldots,k\}$, using $k-1$ documents from a random data point for the other slots. The single relevant document is accompanied by a [RELEVANT] header, while the other documents have an [IRRELEVANT] header, and we prompt the LLM to only summarize the relevant document. For reference-free evaluation metrics, we use the single relevant document as source. We also include a \emph{random} baseline of shuffled inputs and model predictions.
In \Cref{mds_salient}, we see a noticeable drop in performance for all metrics when the salient document is not in the first or final position. Flan-UL2 seems to focus on the \emph{\textbf{end}} of the context, Xgen-7B and Mistral-7B on the \emph{\textbf{beginning}}, and Llama-2 models and GPT-3.5 on \emph{\textbf{both}}. Performance can fall quite below random range, especially for reference-free metrics, confirming the worrying trend from \citet{litm}.

A more fine-grained analysis with GPT-3.5 in \Cref{mds_salient_gpt} evaluating specific attributes (following the method in \citet{cod}, see \Cref{sec:appendix_b}) reveals more details. \emph{Coherence} and \emph{quality} remain high and stable. In other words, the text outputs from the LLMs are always of good quality. But for \emph{informativeness} and \emph{attributability}, the U-shape appears again: it shows that the LLMs (even the powerful GPT-3.5) are struggling to generate content specifically sticking to the document inserted in the middle. 

\begin{table}
\centering
\resizebox{1.0\columnwidth}{!}{
\begin{tabular}{lll|cccccc}
\toprule 

\textbf{Dataset} 
& \textbf{Model} 
& \textbf{Input documents} 
& \textbf{R-2} 
& \textbf{BS} 
& \textbf{A3CU} 
& \textbf{SummaC} 
& \textbf{GPT-3.5} 
& \textbf{\%}\\

\midrule 

\multirow{15}{*}{\textbf{Multi-X}}

& \multirow{3}{*}{Llama-2-7B} 
& All 7                      & 4.64 & 82.83 & 5.88 & 54.56 & 4.17 & 100.00 \\
& & First + last             & 4.62 & 82.82 & 5.78 & 47.00 & 4.50 & 98.36 \\
& & First + 5 random + last  & 4.43 & 82.64 & 5.37 & 43.25 & 4.01 & 92.40 \\

\cdashline{2-9}

& \multirow{3}{*}{Llama-2-13B} 
& All 7                      & 4.78 & 83.00 & 6.64 & 42.62 & 4.35 & 100.00 \\
& & First + last             & 4.73 & 82.86 & 5.76 & 43.74 & 4.53 & 98.46 \\
& & First + 5 random + last  & 4.61 & 82.80 & 5.72 & 46.42 & 4.31 & 98.07 \\

\cdashline{2-9}

& \multirow{3}{*}{Xgen-7B} 
& All 7                      & 5.37 & 82.68 & 6.59 & 44.34 & 4.19 & 100.00 \\
& & First + last             & 5.01 & 82.73 & 5.86 & 49.08 & 4.45 & 99.83 \\
& & First + 5 random + last  & 3.89 & 82.16 & 5.03 & 55.29 & 3.01 & 88.93 \\

\cdashline{2-9}

& \multirow{3}{*}{Mistral-7B} 
& All 7                      & 5.40 & 82.60 & 6.35 & 63.78 & 4.26 & 100.00 \\
& & First + last             & 5.12 & 82.67 & 6.15 & 60.91 & 4.76 & 99.80 \\
& & First + 5 random + last  & 4.45 & 82.40 & 5.35 & 58.19 & 4.06 & 90.59 \\

\cdashline{2-9}

& \multirow{3}{*}{GPT-3.5} 
& All 7                      & 5.26 & 83.45 & 7.71 & 35.66 & 4.59 & 100.00 \\
& & First + last             & 4.75 & 83.04 & 6.05 & 39.81 & 4.69 & 96.42 \\
& & First + 5 random + last  & 4.37 & 82.90 & 5.67 & 43.26 & 4.63 & 95.63 \\

\midrule 

\multirow{15}{*}{\textbf{Multi-N}}   

& \multirow{3}{*}{Llama-2-7B} 
& All 5                      & 10.76 & 85.04 & 19.06 & 60.09 & 4.00 & 100.00 \\
& & First + last             & 9.50 & 84.43 & 15.88 & 54.52 & 3.80 & 91.32 \\
& & First + 3 random + last  & 7.57 & 83.36 & 12.39 & 50.35 & 2.94 & 78.13 \\

\cdashline{2-9}

& \multirow{3}{*}{Llama-2-13B} 
& All 5                      & 10.42 & 84.60 & 18.15 & 57.26 & 3.83 & 100.00 \\
& & First + last             & 9.55 & 84.58 & 16.99 & 49.84 & 3.73 & 93.93 \\
& & First + 3 random + last  & 8.27 & 83.79 & 14.84 & 50.09 & 3.18 & 86.14 \\

\cdashline{2-9}

& \multirow{3}{*}{Xgen-7B} 
& All 5                      & 9.04 & 83.18 & 17.05 & 60.55 & 3.32 & 100.00 \\
& & First + last             & 7.82 & 83.27 & 14.18 & 51.59 & 3.60 & 92.68 \\
& & First + 3 random + last  & 6.30 & 81.85 & 11.66 & 49.02 & 2.66 & 79.51 \\

\cdashline{2-9}

& \multirow{3}{*}{Mistral-7B} 
& All 5                      & 9.52 & 83.55 & 17.03 & 63.02 & 3.15 & 100.00 \\
& & First + last             & 9.11 & 83.69 & 14.99 & 67.14 & 3.51 & 100.37 \\
& & First + 3 random + last  & 6.59 & 81.66 & 12.50 & 52.14 & 2.45 & 80.17 \\

\cdashline{2-9}

& \multirow{3}{*}{GPT-3.5} 
& All 5                      & 10.26 & 85.06 & 18.45 & 49.21 & 4.09 & 100.00 \\
& & First + last             & 8.94 & 84.52 & 15.85 & 45.94 & 4.01 & 92.76 \\
& & First + 3 random + last  & 8.53 & 84.31 & 15.33 & 44.67 & 3.80 & 89.81 \\

\bottomrule

\end{tabular}
}
\caption{\small Performance in multi-document summarization on Multi-XScience (7 documents) and Multi-News (5 documents) when infilling the middle of the context window with random documents. \textbf{R-2} is ROUGE-2, \textbf{BS} refers to BERTScore. \textbf{\%} is the mean relative performance across all metrics compared to the baseline with all documents.}
\label{tab:filling}
\end{table}


\begin{figure*}[t]
    \centering 
    \includegraphics[width=\textwidth]{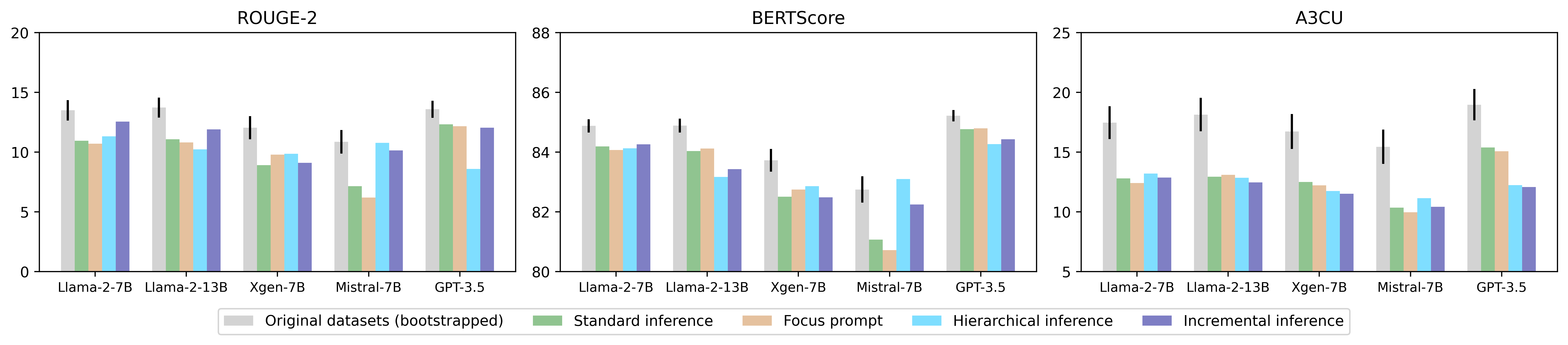}
   \vspace{-2.0em}
    \caption{\small Reference-based evaluation on the MiddleSum dataset. We also report (gray bars) performance achieved by uniformly sampling subsets of the same size as MiddleSum from the original datasets, alongside bootstrapping variance (black lines).}
    \label{middlesum}
\end{figure*}

In the second experiment, we take the opposite approach, and put salient information at the beginning and the end, while the middle of the prompt is filled with noise. We keep the first and last documents, and fill the $k-2$ middle ones with random documents. We also run a baseline just using the first and last documents as input, expected to be close to the result with random documents in between. As displayed in \Cref{tab:filling}, filling with random noise between the first and last document (which amounts to a prompt mostly irrelevant to the reference) leads to a moderate drop in performance. For instance, on Multi-XScience, with 5 random documents between the first and last, Llama-2-13B maintains 98\% of its performance, and reaches a GPT-3.5 score of 4.31 as compared to 4.35 when using all 7 documents.

We conclude from these two experiments that \emph{\textbf{LLMs can focus on the beginning and/or the end of their input, but largely ignore the middle. The U-shape or \emph{middle curse} from \citet{litm} also applies to abstractive summarization.}}

\subsection{Can we alleviate the \emph{middle curse}?}

To evaluate the loss of performance due to the \emph{middle curse} in a natural setup, we subsample data points from each of the 5 long-input summarization datasets. We obtain sentences from the (untruncated) source aligned with the reference summary, following the procedure from \Cref{subsec:rq2}. Only data points where the start index of the earliest aligned source sentence is at least 1,200 words, are kept, ensuring no salient information at the start. 
We randomly sample 50 data points from each of Arxiv, PubMed, GovReport and Multi-News, and 25 from SummScreenFD, forming an evaluation dataset of 225 samples which we name \textbf{MiddleSum}.

We evaluate LLMs on MiddleSum, keeping only reference-based evaluation as the dataset is built using saliency with regard to the reference. As expected, in \Cref{middlesum} we see that LLMs perform noticeably worse on MiddleSum (green bars) as compared to the full set (gray bars), confirming that MiddleSum is a more challenging task.

We benchmark alternative inference methods on MiddleSum: \emph{hierarchical} summarization and \emph{incremental} summarization, both of which are explored in the concurrent work of \citet{booookscore}. Namely, let us divide an input $\vx$ of length $n$ into $k$ consecutive blocks of size at most $m$ (yielding $k = \left \lceil{\frac{n}{m}}\right \rceil $): $\vx = (\vx_{1},\ldots,\vx_{k})$.

\emph{Hierarchical} summarization consists in summarizing each block and then summarizing the concatenation of summaries: 

\begin{align}
    \vy_{i} &= \text{LLM}(\vx_{i}) \quad \forall i \in \{1,\ldots,k\} \\
    \vy &= \text{LLM}(\vy_{1},\ldots,\vy_{k})
\end{align}

\emph{Incremental} summarization consists in updating a summary of the text so far with content from the current text block (we have $\vy_{0}=\emptyset$):

\begin{align}
    \vy_{i} &= \text{LLM}(\vy_{i-1}, \vx_{i}) \forall i \in \{1,\ldots,k\}  \\ 
    \vy &= \vy_{k}
\end{align}


\begin{figure*}
    \centering 
    \includegraphics[width=\textwidth]{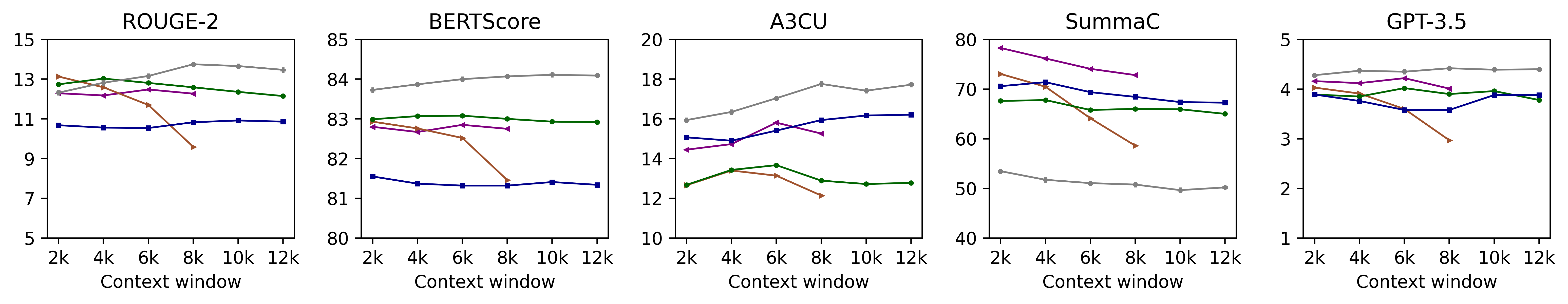}
    \includegraphics[width=\textwidth]{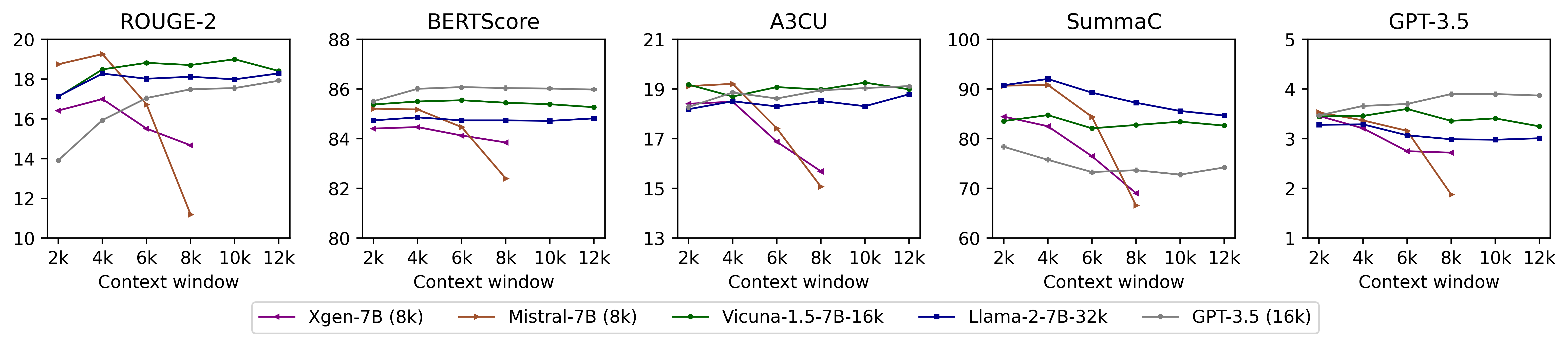}
   \vspace{-2.0em}
    \caption{\small Long-input summarization performance on Arxiv (top) and GovReport (bottom) with 5 LLMs and all 5 metrics. X-axis represents the truncated maximum source length. Xgen-7B and Mistral-7B cannot infer beyond 8k tokens.}
    \label{trunc}
\end{figure*}

In both methods, the final output is $\vy$.
Noting $l$ the output length, standard inference has complexity in $\mathcal{O}(l.n^{2})$, while both alternative methods have complexity in $\mathcal{O}(k.l.m^{2}) = \mathcal{O}(l.n.m)$, which is lower. For both methods and all models, we use a block size $m$ of 1,500 words (roughly 2,000 tokens), and preserve coherence by ending blocks at the earliest end of sentence reaching the length. 

Results are shown in \Cref{middlesum} (blue and purple bars), with detailed numbers in \Cref{sec:appendix_f}. We also compare to a baseline consisting in adding the prompt \texttt{Please also pay attention to the middle section of the input when constructing the summary}, which we refer to as the \emph{Focus prompt} (brown bars). Both alternative methods show promising results on open-source LLMs, notably on Mistral-7B for which they improve performance significantly. However, they are not successful and lag behind \emph{Focus Prompt} with GPT-3.5. Across domains (see \Cref{middlesum_results}), \emph{hierarchical} and \emph{incremental} inference are very effective on scientific publications, which we hypothesize is due to the natural division in structured sections of such inputs. Yet, they seem to harm summaries on the other domains.

\subsection{Is scaling context length \emph{really} useful?}

Experiments from \Cref{subsec:4_1} confirm that LLMs struggle to summarize information contained in the middle of their context window. This poses issues for long-input summarization: after the initial part with (usually) high saliency, important information becomes sparser, and at the same time LLMs processing capability weakens. To investigate this issue, we infer long-document summarization with length truncated at $m*2k$ tokens, varying $m$ from 1 to 6.\footnote{Our hardware does not allow us to exceed 12k tokens.} We use our longest context LLMs \{Xgen-7B, Mistral-7B, GPT-3.5\} ; as well as two open-source LLMs extending Llama-2-7B context window with \textbf{position interpolation} \citep{posinterp}, a method gaining traction as an efficient way to scale LLMs' context window. 
We use Vicuna-7B-1.5-16k\footnote{In HuggingFace: \emph{lmsys/vicuna-7b-v1.5-16k}}, and Llama-2-7B-32k\footnote{In HuggingFace: \emph{togethercomputer/LLaMA-2-7B-32K}}, with context of 16k tokens and 32k tokens, respectively. 

Results on Arxiv and GovReport in \Cref{trunc} confirm our intuition: all metrics plateau or even \emph{decrease} (see Mistral-7B) from 4k context window upwards. Two conflicting forces are at play when increasing length: giving more information to the model helps it retrieve key elements further to make a richer summary, while at the same time reasoning over a longer context is more challenging. 
Yet, such a drop for Xgen-7B and Mistral-7B at 8k inference length is concerning. Both position interpolated models show more robustness ; while GPT-3.5 seems to plateau at 8k tokens.
Our results suggest that \emph{\textbf{in the current LLMs inference and evaluation framework, there is no need to exceed 4k tokens in the context window for open-source model.}} 

\subsection{Does the decoding method impact the \emph{middle curse}?}

We now turn our attention to the process controlling summary generation. While we had sampled all summaries with \textbf{top-k} sampling with $k=50$ and temperature $T=0.3$ so far ; we now also experiment with \textbf{greedy} decoding, and \textbf{top-p} sampling (where we use $p=0.95$ and temperature $T=1.0$). 


\begin{figure}
    \centering 
    \includegraphics[width=\columnwidth]{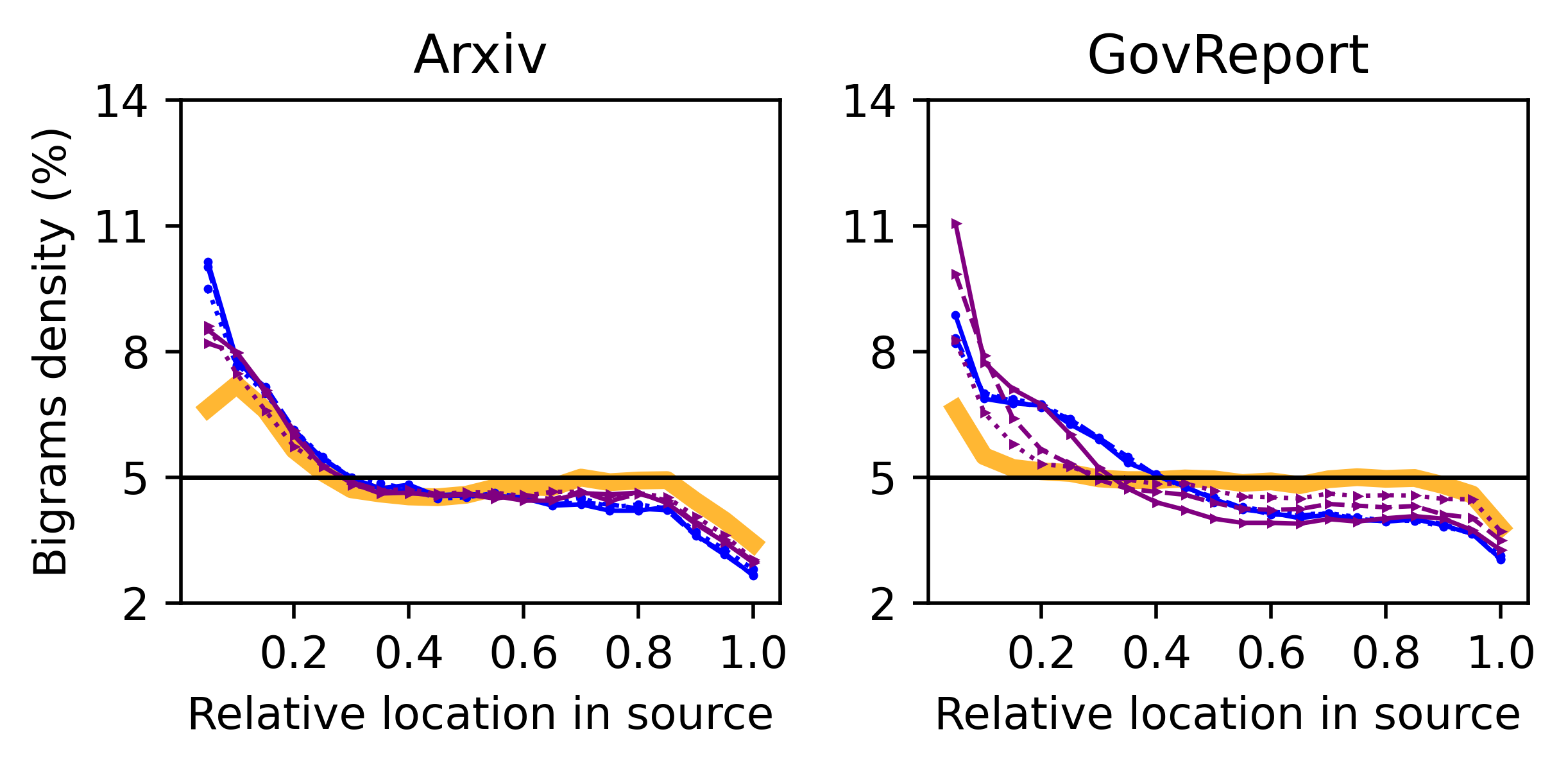}
    \includegraphics[width=\columnwidth]{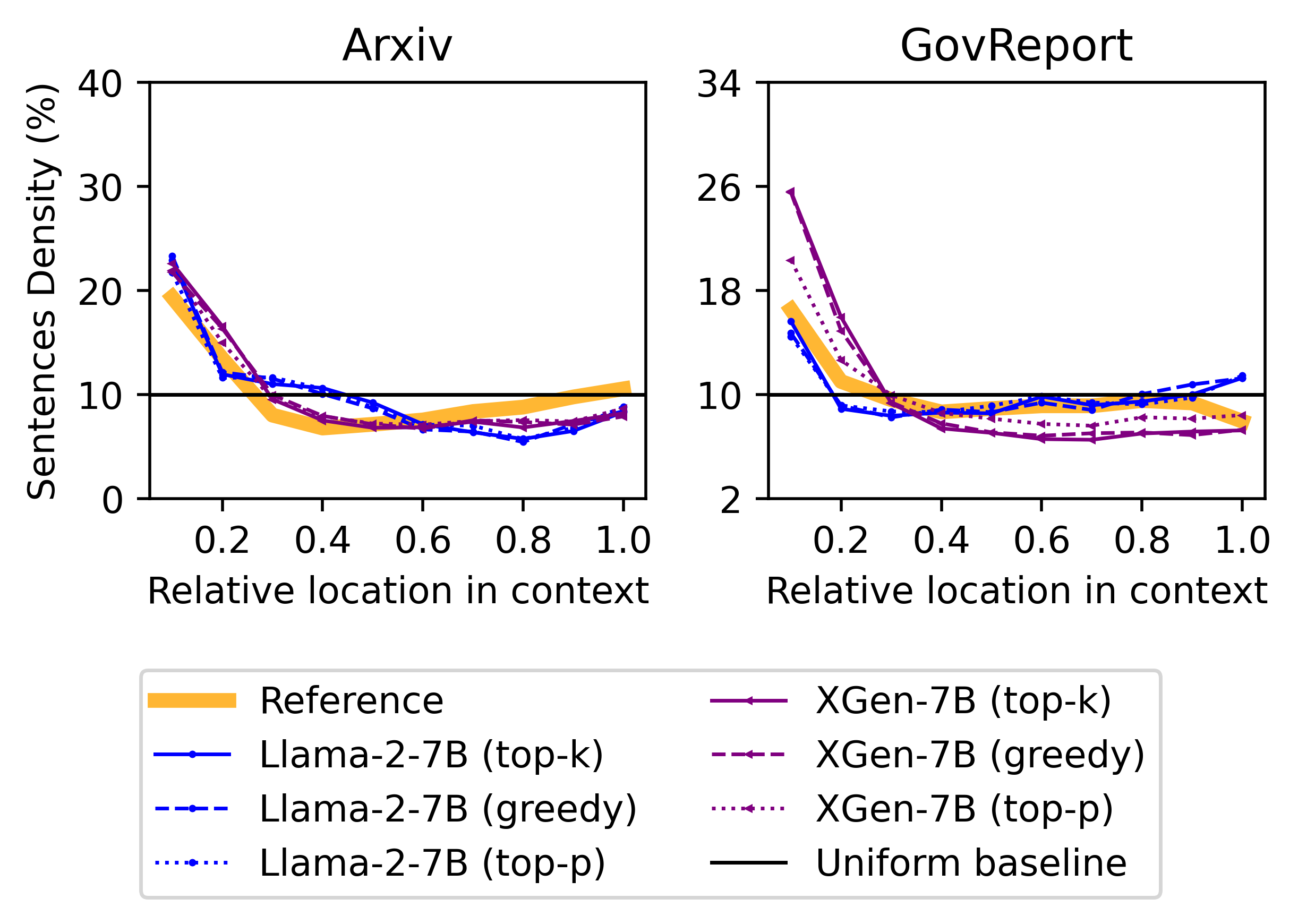}
    \vspace{-1.0em}
    \caption{\small Summary bigrams (top) and aligned source sentences (bottom) distribution on Arxiv and GovReport for Llama-2-7B and XGen-7B, for several decoding strategies.}
    \vspace{-1.0em}
    \label{sampling}
\end{figure}

In \Cref{sampling}, we reproduce the salient bigrams and sentences alignment experiments from \Cref{subsec:rq1} and \Cref{subsec:rq2}, respectively, with the aforementioned decoding methods. As seen, the decoding method does not affect position bias: for all setups, the LLMs show similar patterns as with our previous default decoding method. We conclude that \emph{\textbf{the middle curse is independent from the decoding method}}.

\section{Related Work}
\label{sec:rw}
\paragraph{Summarization with LLMs} 
It is widely acknowledged that LLMs have propelled forward abstractive summarization research \citep{pu2023summarization}, with their summaries being highly rated by human annotators \citep{goyal2022news, zhang2023benchmarking}. \citet{liu2023learning} proposes to train smaller models like BART \citep{bart} or BRIO \citep{brio} with contrastive learning using LLMs like ChatGPT as evaluator providing signal on which generated summary candidate is better. Summary chain-of-thought designs a custom chain-of-thought method which first prompts the LLM to list important facts, then integrates these facts into a coherent summary \citep{wang-etal-2023-element}. SummIt utilizes ChatGPT to iteratively write then refine summaries given feedback from an evaluator LLM \citep{zhang2023summit}. Chain-of-density gradually makes GPT-4 generated summaries contain more and more entities while keeping length budget constant, creating more informative albeit a bit less readable summaries \citep{cod}. \citet{summafusion} noticed that data points with higher compression are generally harder to summarize with pre-trained models. 

\paragraph{Position bias in LLMs}
\citet{longrange} showed that for Transformer-based models, most recent tokens play a greater role compared to older tokens for next-token prediction. 
It was later found that for in-context learning, the order of examples within the prompt impacts GPT-3's performance \cite{liu2021makes,lu2021fantastically}. 
Reliance on positional information affects LLMs capabilities in arithmetic \citep{shen2023positional}, in multiple-choice question-answering \citep{zheng2023large,pezeshkpour2023large}, and as text generation evaluators \citep{wang2023large} ; making it hard to rank LLMs \citep{alzahrani2024benchmarks}.
\citet{litm} were the first to show that LLMs' performance weakens in the middle of the prompt (the \textbf{middle curse}), yet, how LLMs make use of their full context window remains poorly understood. 
The \emph{passkey retrieval} evaluation, which consists in prompting the LLM to recall a complex string or long number inserted in its prompt, is becoming popular recently as a way of verifying LLM's processing capability at each position \cite{litm, mixtral}. However, this task does not measure position bias on complex, abstract reasoning tasks like summarization. 
A line of work attempts to solve the \emph{middle curse} through compressing the prompt \cite{llmlingua, longllmlingua}, with very promising results albeit at the cost of prompt fluency.
Another approach marginalizes results over different permutations of the input to suppress dependency on input order \cite{tang2023found}.
Concurrent work to ours also finds that in zero-shot summarization, LLMs tend to prefer lead content \cite{chhabra2024revisiting}.

\section{Conclusion}
\label{sec:conclusion}
Behind the recent hype around LLMs and their amazing instruction following and content generation capabilities, our study showcases a major weakness in abstractive summarization: LLMs suffer from the \emph{middle curse} and struggle to use information in the middle of their context window. LLMs do not make a consistent use of their context window as they mostly look at the beginning and (to a lesser extent) the end, which at first glance may be hidden by the prevalent \emph{lead bias} in summarization datasets. Extending context window beyond 4k tokens, which has been an intense area of focus lately, is not justified in the current inference and evaluation setup in abstractive summarization. We benchmarked two alternative inference methods (\emph{hierarchical} inference and \emph{incremental} inference) on MiddleSum, an evaluation subset designed to showcase the \emph{middle curse}. Despite encouraging results on scientific paper datasets, these methods are far from a silver bullet to the \emph{middle curse}. We call for a better evaluation of LLMs, which accounts for the salient spans of the source which are effectively being processed.

\section*{Limitations}

Our work mainly considers open-source LLMs for summary generation and ignores some popular closed-source LLMs such as Anthropic's Claude. We made this decision to advocate for openness in LLM research ; yet we acknowledge that it would be interesting to also investigate properties of summaries generated by more paying API LLMs.

Another limitation lays in the saliency estimation. We approximate salient content in the source through maximizing ROUGE-1 overlap with summary sentences. Other methods are also well-suited for this task, albeit at greater computational cost ; for instance semantic similarity through BERTScore or BARTScore ; or saliency estimation through a LLM in zero-shot. 

Lastly, we can only evaluate a finite number of LLMs, and we settled for the evaluation of 6 recent and popular LLMs. Findings may change as LLMs undergo changes and improvements in their pre-training and instruction-tuning methodologies.

\section*{Acknowledgements}

This research was supported by the SINGA scholarship and partially supported by the National Research Foundation, Prime Minister’s Office, Singapore under its Campus for Research Excellence and Technological Enterprise (CREATE) programme. Any opinions, findings, conclusions, or recommendations expressed in this material are those of the author(s) and do not reflect the views of the National Research Foundation, Singapore.

We thank anonymous ARR Reviewers 1 and 3 for their great review, which have led us to include plenty more important results enriching our analysis. We also thank the AC for their constructive meta-review. We thank Florian Le Bronnec and Ruochen Zhao for their helpful proof-reading of the draft.

\bibliography{anthology,custom}
\bibliographystyle{acl_natbib}

\appendix

\section{Statistics}
\label{sec:appendix_a}
\begin{table*}
\resizebox{\textwidth}{!}{
\begin{tabular}{llcccccccccccc}
\toprule
\multirow{2}{*}{\textbf{Dataset}} 
& \multirow{2}{*}{\textbf{Domain}} 
& \multirow{2}{*}{\textbf{\# Docs}} 
& \multicolumn{3}{c}{\textbf{\# Data points}} 
& \multicolumn{3}{c}{\textbf{\# Sentences}} 
& \multicolumn{2}{c}{\textbf{\# Words}} 
& \multicolumn{3}{c}{\textbf{\# Tokens}} \\

\cmidrule(lr){4-6}
\cmidrule(lr){7-9}
\cmidrule(lr){10-11}
\cmidrule(lr){12-14}

& & & Train & Val & Test & Doc. & Summ. & Inf. & Doc. & Summ. & Doc. & Summ. & Max gen.\\

\midrule

CNN/DM \citep{cnndm} & News & 1.00 & 287,113 & 13,334 & \underline{11,490} & 33.37 & 3.79 & 3 & 773.23 & 57.75 & 994.56 & 84.47 & 192 \\
XSum \citep{xsum} & News & 1.00 & 204,045 & 11,332 & \underline{11,334} & 19.11 & 1.00 & 1 & 433.05 & 23.19 & 566.79 & 31.63 & 64 \\
Reddit-TIFU (Long) \citep{reddit} & Social Media & 1.00 & 33,701 & 4,214 & \underline{4,221} & 22.21 & 1.45 & 2 & 444.20 & 23.37 & 532.18 & 29.82 & 128 \\
SAMSum \cite{samsum} & Dialogue & 1.00 & 14,732 & 818 & 819 & 8.96 & 2.01 & 2 & 126.93 & 23.12 & 175.54 & 29.69 & 128 \\
Multi-XScience \citep{multixscience} & Science & 5.06 & 30,369 & 5,066 & \underline{5,093} & 30.55 & 4.86 & 5 & 773.36 & 120.65 & 965.99 & 157.77 & 384 \\ 

\cdashline{1-14}

Arxiv \citep{arxivpubmed} & Science & 1.00 & 203,037 & 6,436 & \underline{6,440} & 250.37 & 6.23 & 6 & 6,446.11 & 166.72 & 8,940.00 & 225.58 & 512 \\ 
PubMed \citep{arxivpubmed} & Science (medical) & 1.00 & 119,924 & 6,633 & \underline{6,658} & 101.61 & 7.59 & 7 & 3,142.92 & 208.03 & 4,602.62 & 324.97 & 512 \\ 
GovReport \citep{govreport} & Legal & 1.00 & 17,517 & 973 & 973 & 282.86 & 23.14 & 22 & 8,363.22 & 649.01 & 11,025.02 & 879.10 & 768 \\ 
SummScreenFD \citep{summscreen} & TV Transcripts & 1.00 & 3,673 & 338 & 337 & 727.06 & 5.26 & 5 & 7,618.20 & 123.34 & 10,067.39 & 157.44 & 512 \\ 
Multi-News \citep{multinews} & News & 2.73 & 44,972 & 5,622 & \underline{5,622} & 79.02 & 9.88 & 10 & 2,101.49 & 256.55 & 2,998.52 & 324.29 & 512 \\ 

\bottomrule

\end{tabular}
}
\caption{\small Statistics on the 10 datasets used for experiments. \textbf{Doc.} is the source document, \textbf{Summ.} the ground-truth summary, \textbf{Inf.} refers to the number of desired sentences to be in the summary prompted to each LLM during inference. Statistics are computed on the entire test set. \textbf{\# Tokens} is calculated with Llama-2's tokenizer. \textbf{Max gen.} is the maximum tokens size that we set when decoding summaries. \underline{Underlined} test sizes correspond to datasets where we subsample randomly 1,000 test data points for evaluation.}
\label{data}
\end{table*}

\begin{table*}
\resizebox{\textwidth}{!}{
\begin{tabular}{llcccccc|cccccc}

\toprule 

\textbf{Model}                        
& \textbf{Metric} 
& \rotatebox[origin=l]{0}{\textbf{CNN/DM}} 
& \rotatebox[origin=l]{0}{\textbf{XSum}} 
& \rotatebox[origin=l]{0}{\textbf{Reddit-TIFU}} 
& \rotatebox[origin=l]{0}{\textbf{SAMSum}} 
& \rotatebox[origin=l]{0}{\textbf{Multi-X}} 
& \rotatebox[origin=l]{0}{\textbf{AVG}}
& \rotatebox[origin=l]{0}{\textbf{Arxiv}} 
& \rotatebox[origin=l]{0}{\textbf{PubMed}} 
& \rotatebox[origin=l]{0}{\textbf{GovReport}} 
& \rotatebox[origin=l]{0}{\textbf{SummScreenFD}} 
& \rotatebox[origin=l]{0}{\textbf{Multi-N}} 
& \rotatebox[origin=l]{0}{\textbf{AVG}} \\

\midrule 

\textbf{SOTA}                         
& ROUGE-2 & \emph{24.17} & \emph{27.09} & \emph{11.13} & \emph{29.88} & \emph{4.60} & \emph{19.37} &\emph{21.93} & \emph{23.26} & \emph{30.90} & \emph{10.70} & \emph{13.60} & \emph{20.07} \\

\cdashline{1-14}

\multirow{6}{*}{\textbf{Flan-UL2}}   
& \# sents 
& 2.89 & 1.00 & 1.34 & 2.08 & 2.28 & 1.19 
& 1.78 & 1.27 & 1.80 & 5.49 & 2.79 & 2.63 \\
& ROUGE-2  
& \textbf{20.28} & \textbf{22.74} & \textbf{8.61} & \textbf{28.21} & 3.04 & \textbf{16.58} 
& 9.37 & 7.42 & 4.76 & 4.33 & 7.79 & 6.73 \\
& BERTScore       
& \textbf{88.05} & \textbf{91.94} & \textbf{87.42} & \textbf{92.29} & 81.87 & \textbf{88.31} 
& 83.82 & 83.07 & 83.53 & 84.85 & 84.97 & 84.05 \\
& A3CU     
& \textbf{32.69} & \textbf{32.11} & \textbf{16.89} & \textbf{49.48} & 5.98 & \textbf{27.43} 
& 14.79 & 13.83 & 12.00 & 8.37 & 16.99 & 13.20 \\ 
& SummaC   
& \textbf{69.96} & 24.27 & \textbf{35.76} & \textbf{30.19} & 57.98 & \textbf{43.63} 
& \textbf{67.56} & 60.96 & 73.80 & \textbf{56.00} & \textbf{76.80} & \textbf{67.02} \\ 
& GPT-3.5  
& 3.16 & 3.52 & 1.61 & 2.92 & 3.23 & 2.89
& 3.11 & 3.16 & 2.46 & 2.11 & 3.36 & 2.84 \\

\cdashline{1-14}

\multirow{6}{*}{\textbf{Llama-2-7B}}  
& \# sents 
& 3.00 & 1.27 & 2.00 & 1.83 & 7.77 & 3.17 
& 5.80 & 6.61 & 12.88 & 5.77 & 18.69 & 9.95 \\
& ROUGE-2  
& 14.16 & 7.27 & 4.17 & 15.53 & 4.87 & 9.20 
& \textbf{13.84} & 12.89 & 16.22 & 5.36 & 12.37 & 12.14 \\
& BERTScore       
& 87.32 & 87.47 & 85.87 & 89.95 & 83.32 & 86.79 
& 83.84 & 82.82 & 85.28 & 85.41 & 85.63 & 84.60 \\
& A3CU     
& 29.04 & 14.18 & 12.15 & 35.64 & 6.39 & 19.48 
& 16.78 & 16.60 & 17.23 & 9.66 & 22.23 & 16.50 \\ 
& SummaC   
& 35.58 & 25.24 & 26.38 & 24.56 & 49.39 & 32.23 
& 53.22 & 51.82 & 70.47 & 39.01 & 57.49 & 54.40 \\ 
& GPT-3.5  
& 4.10 & 4.24 & 2.83 & 3.61 & 4.42 & 3.84
& 4.21 & 4.19 & 3.43 & 2.71 & 3.91 & 3.69 \\

\cdashline{1-14}

\multirow{6}{*}{\textbf{Llama-2-13B}}
& \# sents 
& 3.01 & 1.16 & 2.00 & 1.98 & 5.22 & 2.67 
& 5.92 & 7.22 & 27.75 & 5.16 & 12.79 & 11.77 \\
& ROUGE-2  
& 14.10 & 8.61 & 4.22 & 14.16 & 5.29 & 9.28 
& 13.52 & \textbf{15.24} & 17.28 & 5.62 & 12.58 & 12.85 \\ 
& BERTScore       
& 87.40 & 87.94 & 85.85 & 89.45 & \textbf{83.58} & 86.84 
& 83.88 & 84.24 & 85.29 & 85.42 & 85.84 & 84.93 \\
& A3CU     
& 29.57 & 16.30 & 12.94 & 34.19 & 7.29 & 20.06 
& 16.44 & \textbf{19.21} & 17.01 & 10.30 & \textbf{23.09} & 17.21 \\ 
& SummaC   
& 33.83 & 24.07 & 25.76 & 24.81 & 41.70 & 30.03 
& 55.09 & 56.00 & \textbf{76.44} & 38.74 & 53.12 & 55.88 \\ 
& GPT-3.5  
& \textbf{4.12} & 4.34 & 2.91 & 3.45 & 4.45 & 3.85
& 4.06 & 4.13 & 3.66 & 2.69 & 3.89 & 3.69 \\

\cdashline{1-14}

\multirow{6}{*}{\textbf{Xgen-7B}}      
& \# sents 
& 3.93 & 2.24 & 2.62 & 2.34 & 5.94 & 3.41 
& 8.07 & 13.50 & 22.46 & 10.60 & 9.05 & 12.74 \\
& ROUGE-2  
& 14.55 & 6.00 & 3.98 & 14.51 & 5.54 & 8.92 
& 12.31 & 13.99 & 14.68 & 4.24 & 11.07 & 11.26 \\ 
& BERTScore       
& 87.07 & 87.12 & 85.84 & 89.53 & 83.42 & 86.60 
& 83.07 & 82.87 & 83.94 & 83.91 & 84.95 & 83.75 \\
& A3CU     
& 27.88 & 12.75 & 12.56 & 33.18 & 7.45 & 18.76 
& 15.28 & 18.79 & 15.65 & 8.44 & 21.33 & 18.90 \\ 
& SummaC   
& 52.25 & \textbf{37.95} & 28.40 & 25.63 & 44.36 & 37.72 
& 53.28 & 60.56 & 65.22 & 42.03 & 56.45 & 55.51 \\ 
& GPT-3.5  
& 3.82 & 3.97 & 2.78 & 3.52 & 4.37 & 3.69
& 3.96 & 3.99 & 2.78 & 2.42 & 3.58 & 3.35 \\

\cdashline{1-14}

\multirow{6}{*}{\textbf{Mistral-7B}}      
& \# sents 
& 3.10 & 1.12 & 2.64 & 2.25 & 7.73 & 3.37 
& 12.00 & 11.88 & 25.83 & 27.13 & 16.43 & 18.65 \\
& ROUGE-2  
& 16.37 & 7.05 & 4.34 & 14.66 & \textbf{5.57} & 9.60 
& 9.77 & 14.32 & 11.36 & 3.11 & \textbf{12.61} & 10.23 \\ 
& BERTScore       
& 87.50 & 87.45 & 85.71 & 89.78 & 83.16 & 86.72 
& 81.44 & 82.85 & 82.43 & 81.46 & 85.07 & 82.65 \\
& A3CU     
& 30.60 & 13.14 & 13.08 & 32.21 & 6.96 & 19.20 
& 12.66 & 16.31 & 14.92 & 8.40 & 22.22 & 14.90 \\ 
& SummaC   
& 53.67 & 24.79 & 30.51 & 26.82 & \textbf{62.49} & 39.66 
& 57.81 & \textbf{69.03} & 67.07 & 35.76 & 68.50 & 59.63 \\ 
& GPT-3.5  
& 3.92 & 4.30 & 2.73 & 3.63 & 4.47 & 3.81
& 2.83 & 3.63 & 2.01 & 1.88 & 3.60 & 2.79 \\

\cdashline{1-14}

\multirow{6}{*}{\textbf{GPT-3.5}}      
& \# sents 
& 3.00 & 1.00 & 2.00 & 1.98 & 4.99 & 2.59 
& 5.60 & 6.46 & 15.82 & 5.01 & 9.27 & 8.43 \\
& ROUGE-2  
& 13.17 & 8.19 & 5.17 & 14.47 & 5.37 & 9.27 
& 13.78 & 13.90 & \textbf{17.94} & \textbf{6.55} & 12.28 & \textbf{12.89} \\ 
& BERTScore       
& 87.26 & 87.90 & 86.38 & 89.79 & 83.82 & 87.03 
& \textbf{84.14} & \textbf{84.29} & \textbf{85.99} & \textbf{86.13} & \textbf{86.02} & \textbf{85.31} \\
& A3CU     
& 26.72 & 15.21 & 13.37 & 35.85 & \textbf{7.90} & 19.81 
& \textbf{17.98} & 18.76 & \textbf{19.32} & \textbf{12.69} & 22.95 & \textbf{18.34} \\ 
& SummaC   
& 35.14 & 23.22 & 26.00 & 24.92 & 37.56 & 29.37 
& 50.11 & 47.36 & 73.23 & 40.58 & 48.91 & 52.04 \\ 
& GPT-3.5  
& \textbf{4.12} & \textbf{4.58} & \textbf{3.08} & \textbf{3.81} & \textbf{4.61} & \textbf{4.04} 
& \textbf{4.41} & \textbf{4.36} & \textbf{4.03} & \textbf{3.39} & \textbf{4.14} & \textbf{4.07} \\

\bottomrule

\end{tabular}
}
\caption{\small Performance achieved by the LLMs on each dataset for all 5 metrics. \textbf{\# sents} is the average number of sentences in generated summaries. \textbf{Multi-X} is short for Multi-XScience, \textbf{Multi-N} is Multi-News dataset, \textbf{AVG} columns represent the average over standard-length and long-input datasets, respectively. SOTA numbers are taken from \citep{simmcs} on CNN/DM, from \citep{slic} XSum, Reddit-TIFU and SAMSum, from \citep{topdown} for Arxiv and PubMed, from \citep{bartls} for GovReport and SummScreenFD, from \citep{primera} for Multi-News and Multi-XScience. Due to a lack of reported results for other metrics, we only include ROUGE-2 scores for SOTA models. Best scores (outside of SOTA) are in bold.}
\label{performance}
\end{table*}

In \Cref{data}, we include statistics on each of the abstractive summarization datasets under consideration. We use the non-anonymized version for CNN/DM \citep{get}. For Reddit-TIFU, we use the Long subset, and SummScreenFD, is the ForeverDreaming (FD) subset of SummScreen. GovReport and SummScreenFD are part of the long-input benchmarks Scrolls \citep{scrolls} and ZeroScrolls \citep{zeroscrolls}. 

\begin{figure*}
    \centering 
    \includegraphics[width=1.0\textwidth]{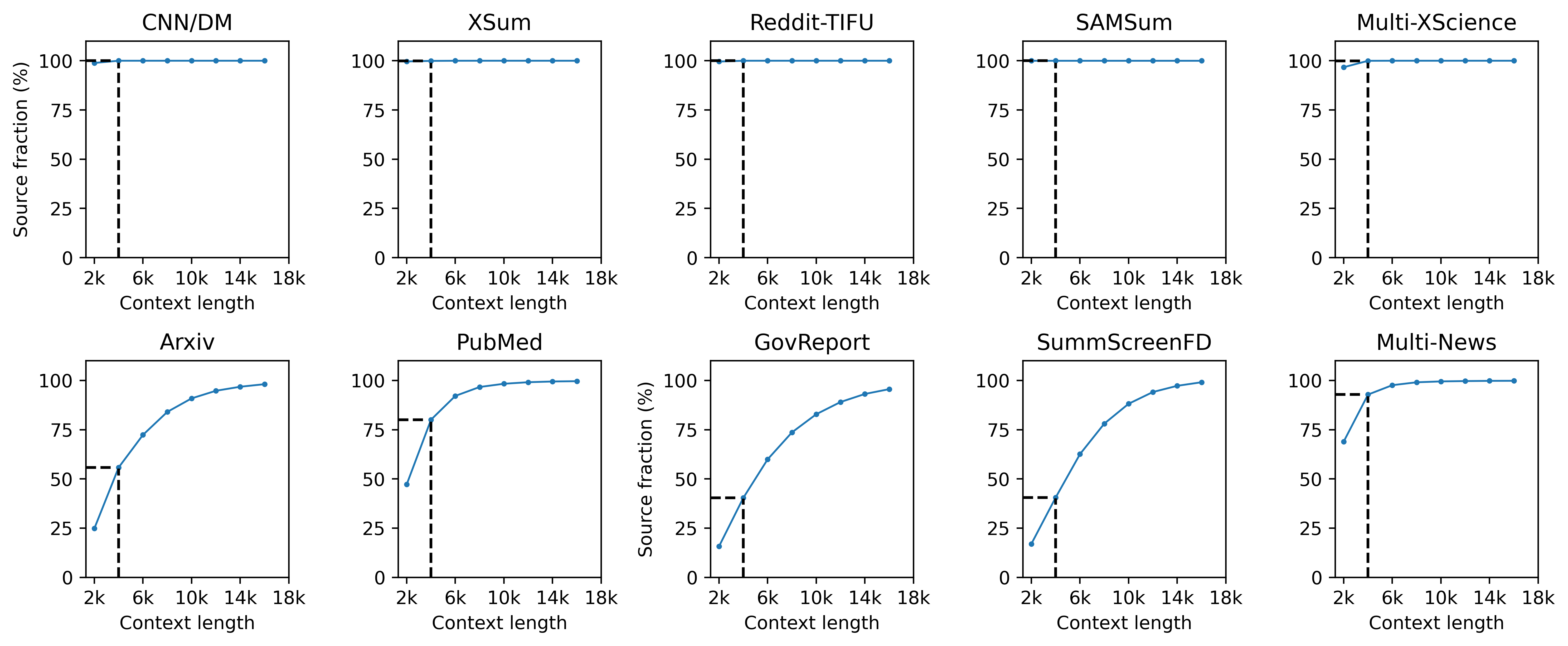}
    \caption{\small Fraction of the source which fits into the context window, for several context window lengths with Llama-2 tokenization. The black dashed lines correspond to Llama-2 context window length of 4k tokens. On standard-length datasets, 4k is enough to access 100\% of all source documents ; but on the long-input datasets such as GovReport or SummScreenFD, such a context window may not even fit 50\% of the source.}
    \label{trunc_frac}
\end{figure*}

In \Cref{trunc_frac}, we illustrate how much of the source document(s) is visible with a 4k context window (Llama-2).


\section{GPT-3.5 Evaluation}
\label{sec:appendix_b}
To evaluate LLM-generated summaries with GPT-3.5, we use the following prompt template: \\

\textit{Score the following summary generated by another system given the source on a scale from 1 to 5 with regards to overall general summary quality. 1-point indicates a low quality summary, and 5 points a very high quality summary. A high quality summary is grammatical, fluent, informative, relevant, coherent and factually consistent with the source. Let's think step-by-step and just output the score.} \\

\textit{Source:}

\textit{[source]}\\

\textit{Instruction:}

\textit{Summarize the above text in [n] sentences.}\\

\textit{Summary:}

\textit{[summary generated by the LLM to evaluate]}\\

\textit{Your score:}\\

When evaluating for the specific aspect of \emph{\textbf{informativeness}}, the first paragraph becomes:\\

\textit{Score the following summary generated by another system given the source on a scale from 1 to 5 with regards to how informative the summary is. 1 point indicates a not informative summary, and 5 points a very informative summary. An informative summary captures the important information in the article and presents it accurately and concisely. Let's think step-by-step and just output the score.}\\

When evaluating for the specific aspect of \emph{\textbf{overall quality}}, the first paragraph becomes:\\

\textit{Score the following summary generated by another system given the source on a scale from 1 to 5 with regards to its quality. 1 point indicates a low quality summary, and 5 points a very high quality summary. A high quality summary is comprehensible and understandable. Let's think step-by-step and just output the score.}\\

\begin{table*}[h]
\resizebox{\textwidth}{!}{
\begin{tabular}{lcccccccccc}

\toprule

\textbf{Model}       
& \textbf{CNN/DM} 
& \textbf{XSum} 
& \textbf{Reddit-TIFU} 
& \textbf{SAMSum} 
& \textbf{Multi-XScience} 
& \textbf{Arxiv} 
& \textbf{PubMed} 
& \textbf{GovReport} 
& \textbf{SummScreenFD} 
& \textbf{Multi-News} \\

\midrule 

\textbf{Flan-UL2}    & 0.000 & \transparent{0.5}0.012 & 0.000 & \transparent{0.5}0.057 & 0.000 & \_ & \_ & \_ & \_ & \_ \\
\textbf{Llama-2-7B}  & 0.000 & 0.000 & 0.000 & \transparent{0.5}0.490 & 0.000 & 0.000 & 0.000 & 0.000 & 0.000 & 0.000 \\
\textbf{Llama-2-13B} & 0.000 & 0.000 & 0.000 & 0.000 & 0.000 & 0.000 & 0.000 & 0.000 & 0.000 & 0.000 \\
\textbf{Xgen-7B}     & 0.000 & 0.000 & 0.000 & 0.000 & 0.000 & 0.000 & 0.000 & 0.000 & 0.000 & 0.000 \\
\textbf{Mistral-7B}  & 0.000 & 0.000 & 0.000 & 0.000 & 0.000 & 0.000 & 0.000 & 0.000 & 0.000 & 0.000 \\     
\textbf{GPT-3.5}     & 0.000 & 0.000 & 0.000 & \transparent{0.5}0.135 & 0.000 & 0.000 & 0.000 & 0.000 & 0.000 & 0.000 \\    

\bottomrule

\end{tabular}
}
\caption{\small P-value of a 2-sample Kolmogorov-Smirnov test between the position distribution of bigrams in LLM-generated summaries and bigrams in reference summaries. We round numbers to 3 decimals. Numbers in gray correspond to non-significant differences (p-value above 0.01).}
\label{ks}
\end{table*}

\begin{table*}[h]
\resizebox{\textwidth}{!}{
\begin{tabular}{llcccccccc}

\toprule 

\multirow{2}{*}{\textbf{Dataset}} 
& \multirow{2}{*}{\textbf{Domain}} 
& \multirow{2}{*}{\textbf{\# Docs}} 
& \multirow{2}{*}{\textbf{\# Data points}} 
& \multicolumn{2}{c}{\textbf{\# Sentences}}   
& \multicolumn{2}{c}{\textbf{\# Words}}       
& \multicolumn{2}{c}{\textbf{\# Tokens}} \\

\cmidrule(lr){5-6}
\cmidrule(lr){7-8}
\cmidrule(lr){9-10}

& & & & \textbf{Doc.} & \textbf{Summ.} & \textbf{Doc.} & \textbf{Summ.} & \textbf{Doc.} & \textbf{Summ.} \\

\midrule

Arxiv  & Science & 1.00 & 50 & 299.36 & 5.84 & 7,605.52 & 165.80 & 10,846.00 & 226.22 \\
PubMed & Science (medical) & 1.00 & 50 & 157.60 & 6.96 & 5,090.84 & 203.04 & 7,783.02 & 329.12 \\
GovReport & Legal & 1.00 & 50 & 445.64 & 22.80 & 13,308.70 & 656.10 & 17,462.90 & 883.10 \\
SummScreen & TV transcripts & 1.00 & 25 & 762.24 & 3.56 & 9,732.44 & 88.60 & 12,878.08 & 115.00 \\
MultiNews & News & 3.18 & 50 & 211.34 & 9.86 & 5,939.32 & 269.84 & 8,130.78 & 336.16 \\

\cdashline{1-10}

Overall & Mixed & 1.48 & 225 & 332.24 & 10.50 & 8,180.13 & 297.57 & 11,258.16 & 407.15 \\

\bottomrule

\end{tabular}
}
\caption{\small Statistics on the MiddleSum evaluation dataset, breaking down on each domain. \textbf{Doc.} is the source document, \textbf{Summ.} the ground-truth summary. \textbf{\# Tokens} is calculated with Llama-2's tokenizer.}
\label{middlesum_stats}
\end{table*}

When evaluating for the specific aspect of \emph{\textbf{coherence}}, the first paragraph becomes:\\

\textit{Score the following summary generated by another system given the source on a scale from 1 to 5 with regards to its coherence. 1 point indicates an incoherent summary, and 5 points a very coherent summary. A coherent summary is well-structured and well-organized. Let's think step-by-step and just output the score.}\\

When evaluating for the specific aspect of \emph{\textbf{attributability}}, the first paragraph becomes:\\

\textit{Score the following summary generated by another system given the source on a scale from 1 to 5 with regards to how attributable it is. 1 point indicates a not very attributable summary, with many hallucinations, and 5 points a summary very attributable to the source, consistent with the source. In a very attributable summary, all the information is fully attributable to the source. Let's think step-by-step and just output the score.}

\section{Baseline Performance}
\label{sec:appendix_c}
\begin{figure}
    \centering 
    \includegraphics[width=1.0\columnwidth]{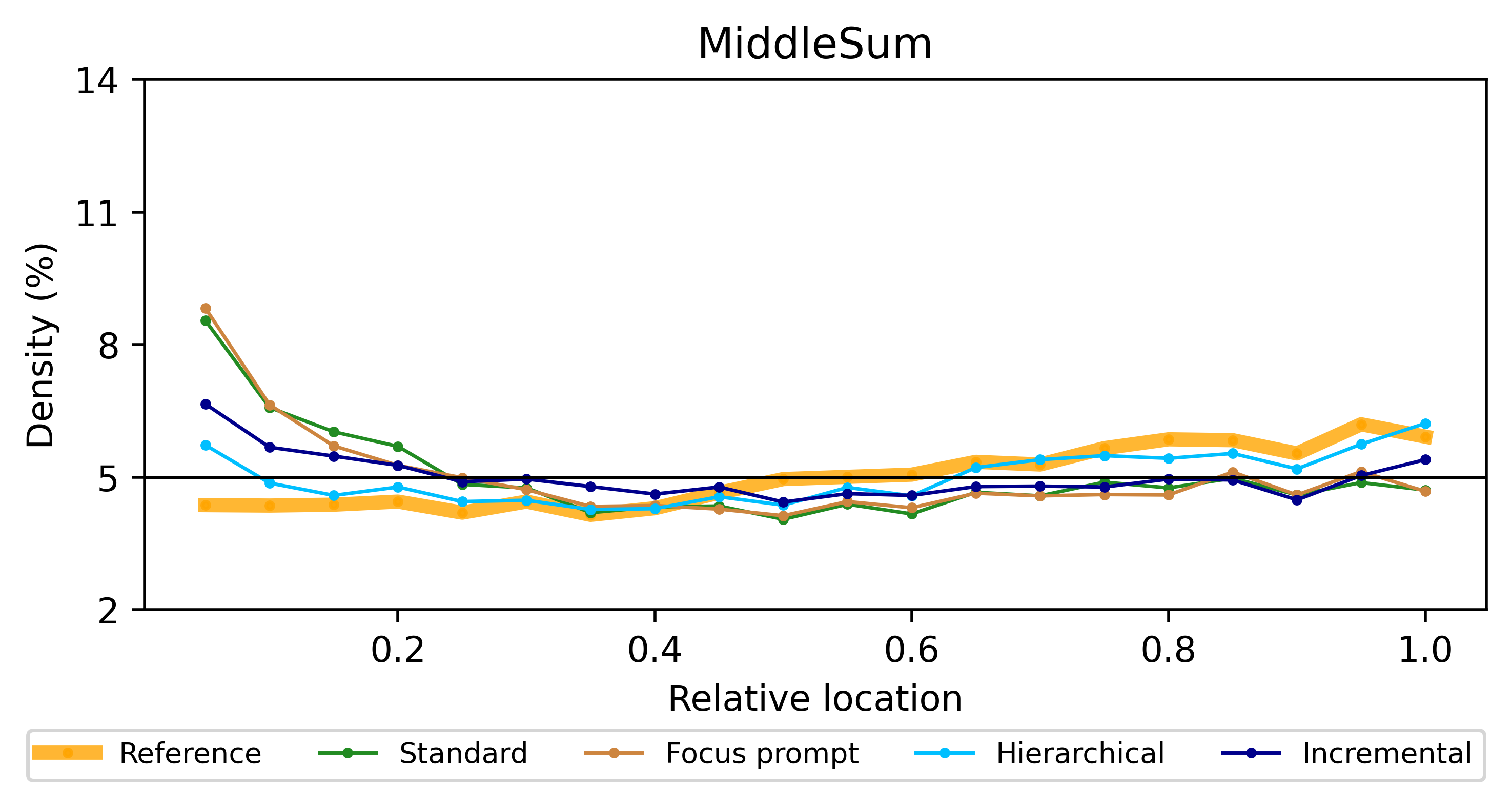}
    \caption{\small Relative location of summary bigrams within the source on MiddleSum for Llama-2-13B with different inference methods.}
    \label{middlesum_bigrams}
\end{figure}

In \Cref{performance}, we report zero-shot performance with the 6 LLMs described in \Cref{sec:setup}. We note that for standard-length datasets, Flan-UL2 is dominating, perhaps due to its better instruction-tuning ; while for long-context ones, GPT-3.5 takes the lead. GPT-3.5 is always consistently ahead on the GPT-3.5 metric, showing the preference of the LLM for its own generation when used as an evaluator.

\section{Statistical Significance on RQ1}
\label{sec:appendix_d}
In \Cref{ks}, we run a 2-sample Kolmogorov-Smirnov statistical significance test to compare the bigrams position distribution of LLMs with the reference summaries.

\section{Results with \emph{base} Models}
\label{sec:appendix_e}

\begin{figure*}[h]
    \centering 
    \includegraphics[width=\textwidth]{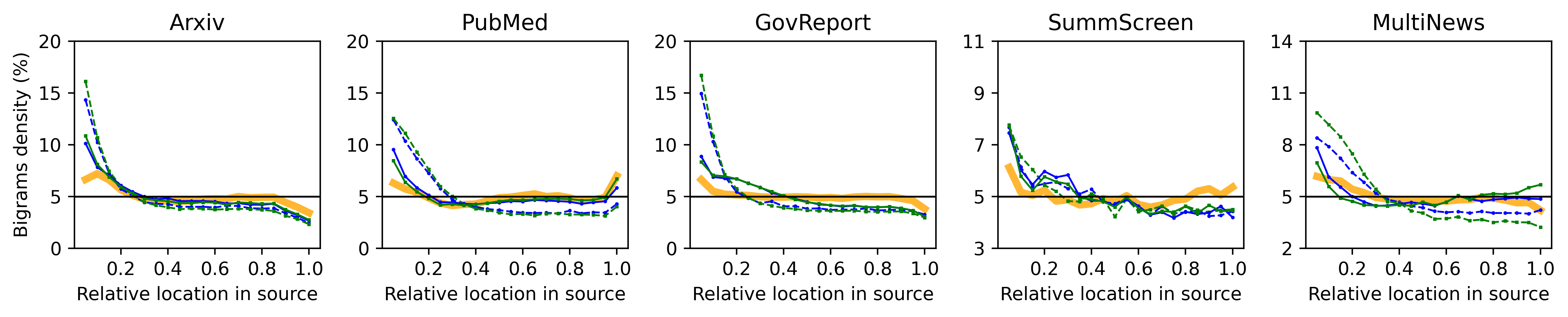}
    \includegraphics[width=\textwidth]{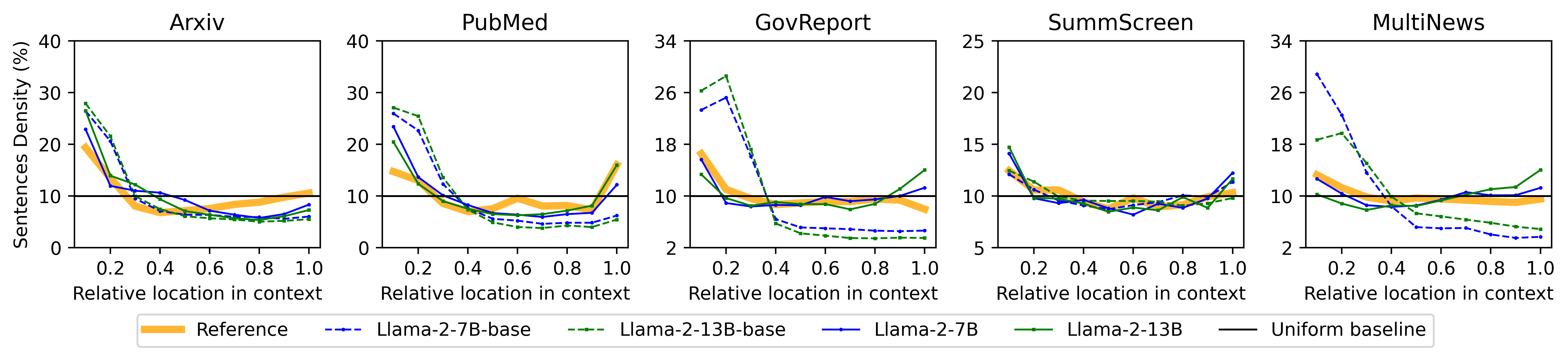}
    \caption{\small Summary bigrams (top) and aligned source sentences (bottom) distribution for Llama-2-7B and Llama-2-13B on the long-input datasets, both with \emph{base} (dashed lines) and \emph{chat} (full lines) models.}
    \label{base}
\end{figure*}

In \Cref{base}, we repeat the analysis from \Cref{subsec:rq1} (bigrams alignment) and \Cref{subsec:rq2} (sentences alignment) but this time using the \emph{base} model versions (not instruction-tuned) of Llama-2-7B and Llama-2-13B. As we can see, the position bias is even stronger for base models than for chatbot models, suggesting that this bias is acquired during pre-training. 

\section{More Results on MiddleSum}
\label{sec:appendix_f}
In \Cref{middlesum_stats}, we show statistics on the MiddleSum evaluation dataset. 

In \Cref{middlesum_bigrams}, we repeat the salient bigrams analysis from \Cref{alignment_absolute} for Llama-2-13B on MiddleSum. We note that both \emph{hierarchical} and \emph{incremental} inference notably decrease reliance on lead bigrams compared to standard inference, while the simple \emph{Focus Prompt} does not. 

In \Cref{middlesum_results}, we show reference-based evaluation on each of the 5 subsets of MiddleSum. 

\begin{table*}[t]
\resizebox{\textwidth}{!}{
\begin{tabular}{lllcccccc}

\toprule 

\textbf{Model}              
& \textbf{Metric}            
& \textbf{Inference} 
& \textbf{MiddleSum (MS)} 
& \textbf{MS/Arxiv} 
& \textbf{MS/PubMed} 
& \textbf{MS/GovReport} 
& \textbf{MS/SummScreen} 
& \textbf{MS/Multi-News} \\

\midrule 

\multirow{12}{*}{Llama-2-7B} 
& \multirow{4}{*}{ROUGE-2}   
& Standard       & 10.96 & 12.62 & 10.97 & 13.26 & 4.07 & \textbf{10.43} \\
& & Focus prompt & 10.70 & 12.10 & 11.51 & 13.28 & 3.57 & 9.48 \\ 
& & Hierarchical & 11.33 & \textbf{14.63} & 13.36 & 13.51 & 4.85 & 7.06 \\
& & Incremental & \textbf{12.56} & 14.43 & \textbf{13.54} & \textbf{17.26} & \textbf{5.19} & 8.68 \\

\cdashline{2-9}

& \multirow{4}{*}{BERTScore}   
& Standard       & 84.19 & 83.43 & 83.13 & 84.63 & 85.33 & \textbf{85.01} \\
& & Focus prompt & 84.07 & 83.33 & 83.00 & \textbf{84.77} & 85.27 & 84.59 \\ 
& & Hierarchical & 84.13 & \textbf{83.86} & 83.74 & 84.26 & 85.23 & 84.10 \\
& & Incremental & \textbf{84.26} & 83.73 & \textbf{83.80} & 84.67 & \textbf{85.70} & 84.14 \\     

\cdashline{2-9}

& \multirow{4}{*}{A3CU}   
& Standard       & 12.81 & 13.55 & 12.12 & 11.71 & 7.31 & \textbf{16.61} \\
& & Focus prompt & 12.41 & 13.54 & 13.39 & 10.84 & 7.15 & 14.52 \\ 
& & Hierarchical & \textbf{13.21} & \textbf{15.64} & \textbf{15.14} & \textbf{12.26} & 9.71 & 11.57 \\
& & Incremental & 12.88 & 15.34 & 14.45 & 10.51 & \textbf{10.24} & 12.55 \\

\midrule 

\multirow{12}{*}{Llama-2-13B} 
& \multirow{4}{*}{ROUGE-2}   
& Standard       & 11.07 & 11.68 & 11.63 & 13.56 & \textbf{5.09} & 10.38 \\
& & Focus prompt & 10.82 & 11.51 & 11.30 & 12.42 & 4.83 & \textbf{11.03} \\ 
& & Hierarchical & 10.24 & \textbf{13.45} & 13.26 & 10.21 & 4.93 & 6.71 \\
& & Incremental & \textbf{11.90} & 12.53 & \textbf{13.84} & \textbf{17.34} & 5.06 & 7.33 \\

\cdashline{2-9}

& \multirow{4}{*}{BERTScore}   
& Standard       & 84.04 & 83.29 & 83.15 & 84.43 & \textbf{85.24} & 84.70 \\
& & Focus prompt & \textbf{84.12} & 83.26 & 83.06 & 84.49 & 85.08 & \textbf{85.19} \\ 
& & Hierarchical & 83.17 & \textbf{83.74} & \textbf{83.75} & 80.07 & 85.03 & 84.20 \\
& & Incremental & 83.43 & 82.45 & 83.45 & \textbf{84.96} & 83.24 & 82.95 \\    

\cdashline{2-9}

& \multirow{4}{*}{A3CU}   
& Standard       & 12.94 & 13.01 & 13.42 & 10.92 & 8.47 & 16.65 \\
& & Focus prompt & \textbf{13.10} & 13.17 & 11.76 & 11.91 & 7.06 & \textbf{18.58} \\ 
& & Hierarchical & 12.85 & \textbf{16.28} & 14.11 & 10.40 & 8.36 & 12.86 \\
& & Incremental & 12.47 & 14.64 & \textbf{14.80} & \textbf{11.75} & \textbf{10.76} & 9.54 \\

\midrule 

\multirow{12}{*}{Xgen-7B} 
& \multirow{4}{*}{ROUGE-2}   
& Standard       & 8.92 & 11.94 & 9.34 & 10.02 & \textbf{4.53} & 6.57 \\
& & Focus prompt & 9.79 & 12.87 & 10.79 & 11.46 & 4.15 & \textbf{6.85} \\ 
& & Hierarchical & \textbf{9.87} & \textbf{13.06} & \textbf{11.37} & 11.77 & 2.79 & 6.83 \\
& & Incremental & 9.11 & 10.97 & 10.55 & \textbf{13.16} & 3.71 & 4.44 \\

\cdashline{2-9}

& \multirow{4}{*}{BERTScore}  
& Standard       & 82.51 & 82.45 & 81.96 & 82.65 & 83.28 & 82.60 \\
& & Focus prompt & 82.75 & 82.45 & 82.14 & 83.08 & \textbf{83.81} & 82.80 \\ 
& & Hierarchical & \textbf{82.86} & \textbf{82.73} & \textbf{82.56} & 82.67 & 83.79 & \textbf{83.03} \\
& & Incremental & 82.49 & 81.94 & 81.82 & \textbf{83.11} & 83.39 & 80.83 \\     

\cdashline{2-9}

& \multirow{4}{*}{A3CU}   
& Standard       & \textbf{12.50} & 15.14 & 12.85 & \textbf{11.99} & 8.48 & \textbf{12.05} \\
& & Focus prompt & 12.23 & \textbf{15.82} & 12.04 & 11.57 & 8.20 & 11.50 \\ 
& & Hierarchical & 11.74 & 14.55 & 12.61 & 10.76 & 7.33 & 11.25 \\
& & Incremental & 11.51 & 12.73 & \textbf{13.15} & 9.13 & \textbf{8.97} & 9.69 \\

\midrule 

\multirow{12}{*}{Mistral-7B} 
& \multirow{4}{*}{ROUGE-2}   
& Standard       & 7.16 & 9.28 & 9.56 & 5.86 & 2.19 & 6.41 \\
& & Focus prompt & 6.20 & 7.62 & 8.44 & 4.85 & 1.42 & 6.29 \\ 
& & Hierarchical & \textbf{10.78} & \textbf{11.67} & \textbf{13.71} & \textbf{13.36} & \textbf{4.76} & \textbf{7.39} \\
& & Incremental & 10.15 & 11.36 & 12.66 & 12.89 & 3.46 & 7.02 \\

\cdashline{2-9}

& \multirow{4}{*}{BERTScore}   
& Standard       & 81.07 & 80.98 & 81.19 & 80.41 & 80.10 & 82.20 \\
& & Focus prompt & 80.72 & 80.31 & 80.81 & 79.91 & 80.12 & 82.13 \\ 
& & Hierarchical & \textbf{83.10} & \textbf{82.56} & \textbf{83.00} & \textbf{83.81} & \textbf{83.90} & \textbf{82.62} \\
& & Incremental & 82.25 & 81.92 & 82.39 & 83.67 & 83.14 & 80.56 \\   

\cdashline{2-9}

& \multirow{4}{*}{A3CU}   
& Standard       & 10.35 & 11.55 & 11.46 & 7.94 & \textbf{8.53} & 11.34 \\
& & Focus prompt & 9.96 & 10.13 & 10.65 & 7.81 & 7.59 & \textbf{12.44} \\ 
& & Hierarchical & \textbf{11.15} & 12.15 & \textbf{14.21} & \textbf{10.62} & 7.48 & 9.47 \\
& & Incremental & 10.43 & \textbf{14.25} & 11.00 & 8.90 & 7.44 & 9.05 \\

\midrule 

\multirow{12}{*}{GPT-3.5} 
& \multirow{4}{*}{ROUGE-2}   
& Standard       & \textbf{12.33} & \textbf{12.99} & \textbf{12.78} & 16.16 & 6.05 & \textbf{10.53}  \\
& & Focus prompt & 12.17 & 12.61 & 12.56 & 15.95 & \textbf{6.35} & 10.46 \\ 
& & Hierarchical & 8.59 & 9.58 & 10.27 & 10.30 & 4.33 & 6.36 \\
& & Incremental  & 12.04 & 11.85 & 12.17 & \textbf{18.46} & 5.38 & 9.03\\

\cdashline{2-9}

& \multirow{4}{*}{BERTScore}   
& Standard       & 84.77 & \textbf{83.81} & 83.70 & 85.41 & \textbf{86.32} & \textbf{85.39} \\
& & Focus prompt & \textbf{84.80} & 83.76 & \textbf{83.81} & \textbf{85.50} & 86.28 & \textbf{85.39} \\ 
& & Hierarchical & 84.27 & 83.34 & 83.62 & 85.30 & 85.37 & 84.28 \\
& & Incremental  & 84.43 & 83.48 & 83.74 & 85.41 & 85.76 & 84.44 \\

\cdashline{2-9}

& \multirow{4}{*}{A3CU}   
& Standard       & \textbf{15.40} & \textbf{15.22} & 15.58 & 12.72 & 13.15 & \textbf{19.21} \\
& & Focus prompt & 15.07 & 14.51 & \textbf{15.59} & 12.34 & \textbf{13.63} & 18.54 \\ 
& & Hierarchical & 12.24 & 12.54 & 13.01 & \textbf{13.29} & 9.03 & 11.74 \\
& & Incremental  & 12.08 & 13.59 & 13.24 & 10.90 & 9.05 & 12.08 \\

\bottomrule

\end{tabular}
}
\caption{\small Reference-based results for all models and inference methods on MiddleSum, breaking down by subset. The best number across inference methods is in bold.}
\label{middlesum_results}
\end{table*}

\section{Software}
\label{sec:appendix_g}
We use the following Python libraries, all open source:
\begin{itemize}[leftmargin=*,itemsep=0.1em]
    \item \emph{numpy}, version 1.24.3 

    \item \emph{torch}, version 2.0.1

    \item \emph{scikit-learn}, version 1.0.2

    \item \emph{sentencepiece}, version 0.1.97

    \item \emph{nltk}, version 3.8.1

    \item \emph{spacy}, version 3.6.0

    \item \emph{scipy}, version 1.10.1

    \item \emph{rouge-score}, version 0.1.2

    \item \emph{bert-score}, version 0.3.13

    \item \emph{summac}, version 0.0.03

    \item \emph{tiktoken}, version 0.4.0

    \item \emph{openai}, version 0.28.0

    \item \emph{huggingface-hub}, version 0.17.2

    \item \emph{datasets}, version 2.14.5

    \item \emph{accelerate}, version 0.21.0

    \item \emph{tokenizers}, version 0.14.1

    \item \emph{transformers}, version 4.34.0
    
\end{itemize}

\end{document}